\documentclass[sigconf]{acmart}

\def\eg{\emph{e.g.}} 

\def\ie{\emph{i.e.}}

\def\al{\emph{et al. }}

\graphicspath{{fig/}}
\usepackage{multirow}
\usepackage{subfig}
\usepackage{textcomp,booktabs}
\usepackage{colortbl}
\definecolor{mygray}{gray}{.9}
\usepackage{algorithmic}
\usepackage[ruled, vlined]{algorithm2e}

\SetKwInput{KwInput}{Input}                
\SetKwInput{KwOutput}{Output}

\AtBeginDocument{%
  \providecommand\BibTeX{{%
    \normalfont B\kern-0.5em{\scshape i\kern-0.25em b}\kern-0.8em\TeX}}}

\copyrightyear{2020}
\acmYear{2020}
\setcopyright{acmcopyright}\acmConference[MM '20]{Proceedings of the 28th ACM International Conference on Multimedia}{October 12--16, 2020}{Seattle, WA, USA} \acmBooktitle{Proceedings of the 28th ACM International Conference on Multimedia (MM '20), October 12--16, 2020, Seattle, WA, USA}
\acmPrice{15.00}
\acmDOI{10.1145/3394171.3413978}
\acmISBN{978-1-4503-7988-5/20/10}

\begin{document}
\fancyhead{}

\title{Salvage Reusable Samples from Noisy Data for Robust Learning}

\author{Zeren Sun}
\orcid{0000-0001-6262-5338}
\affiliation{
	\institution{Nanjing University of Science and Technology, Nanjing, China}
}
\email{zerens@njust.edu.cn}

\author{Xian-Sheng Hua}
\affiliation{
	\institution{Alibaba Group, Hangzhou, China}
}
\email{huaxiansheng@gmail.com}

\author{Yazhou Yao}
\authornote{Corresponding author}
\affiliation{
	\institution{Nanjing University of Science and Technology, Nanjing, China}
}
\email{yazhou.yao@njust.edu.cn}

\author{Xiu-Shen Wei}
\affiliation{	
	\institution{Nanjing University of Science and Technology, Nanjing, China}
}
\email{weixs.gm@gmail.com}

\author{Guosheng Hu}
\affiliation{
	\institution{AnyVision, Belfast, United Kingdom}
}
\email{huguosheng100@gmail.com}

\author{Jian Zhang}
\affiliation{
	\institution{University of Technology Sydney, Sydney, Australia}
}
\email{jian.zhang@uts.edu.au}

\renewcommand{\shortauthors}{Sun et al.}

\begin{abstract}

Due to the existence of label noise in web images and the high memorization capacity of deep neural networks, training deep fine-grained (FG) models directly through web images tends to have an inferior recognition ability. In the literature, to alleviate this issue, loss correction methods try to estimate the noise transition matrix, but the inevitable false correction would cause severe accumulated errors. Sample selection methods identify clean (``easy'') samples based on the fact that small losses can alleviate the accumulated errors. However, ``hard'' and mislabeled examples that can both boost the robustness of FG models are also dropped. To this end, we propose a certainty-based reusable sample selection and correction approach, termed as CRSSC, for coping with label noise in training deep FG models with web images. Our key idea is to additionally identify and correct reusable samples, and then leverage them together with clean examples to update the networks. We demonstrate the superiority of the proposed approach from both theoretical and experimental perspectives. 
The source code, models, and data have been made available at \url{https://github.com/NUST-Machine-Intelligence-Laboratory/CRSSC}.

\end{abstract}



\maketitle

\section{Introduction}

In the past few years, labeled image datasets have played a critical role in computer vision tasks \cite{shu2019hierarchical,luo2019segeqa,tang2017personalized,chen2020classification,lu2020hsi}. To distinguish the subtle differences among fine-grained categories (\eg, birds \cite{cub200-2011}, airplanes \cite{fgvc-aircraft}, or plants \cite{inat17}), a large amount of well-labeled images are typically required. However, labeling objects at the subordinate level generally requires domain-specific expert knowledge, which is not always available for a human annotator from crowd-sourcing platforms like Amazon Mechanical Turk \cite{xie2019attentive,xie2020eccv}.

\begin{figure}[t]
\centering
\includegraphics[width=0.9\linewidth]{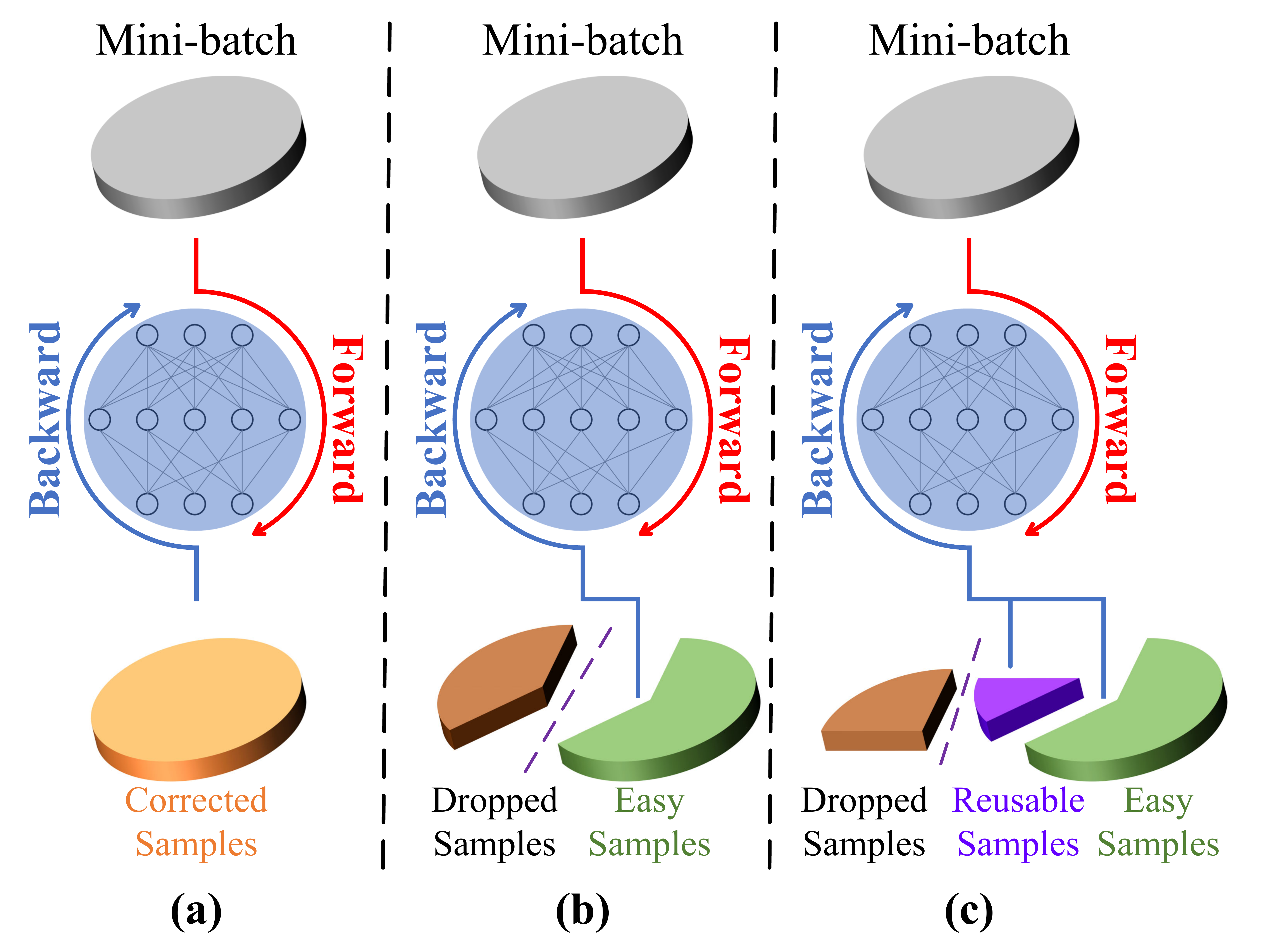}
\caption{\small{(\textbf{a}) Loss correction methods correct losses for all samples in each mini-batch before back-propagating. (\textbf{b}) Sample selection methods identify easy samples out of each mini-batch and then back-propagate using only these easy samples. (\textbf{c}) Our CRSSC is based on sample selection methods, but proposes to leverage additionally reusable samples (including ``\textbf{hard}'' and \textbf{mislabeled} ones) for boosting the model learning.}}
\label{fig1}
\end{figure} 

To reduce the cost of fine-grained annotation, many methods have been proposed which mainly focus on a semi-supervised learning paradigm \cite{cui2016fine,niu2018webly,xu2018webly,yao2017exploiting}. For example, Xu \al \cite{xu2018webly} proposed to utilize detailed annotations and transfer as much knowledge as possible from existing strongly supervised datasets to weakly supervised web images for fine-grained recognition. Niu \al \cite{niu2018webly} proposed a new learning scenario which only required experts to label a few fine-grained subcategories and can predict all the remaining subcategories by virtue of web data. Nevertheless, semi-supervised methods involve various forms of human intervention and have relatively limited scalability.

To further reduce the demand of manual annotation, leveraging web images to train FGVC models has attracted broad attention \cite{aaai20,yao2019towards,yao2018extracting,yao2020exploiting}. Owing to the error-prone automatic tagging system or non-expert annotation, web images for fine-grained categories are usually associated with massive label noise. Therefore, along with how to find discriminative regions in fine-grained images, how to handle label noise is another pivotal problem for training deep FGVC models with web images, which is also the focus of this paper. Statistical learning has contributed significantly to cope with label noise, especially in theoretical aspects \cite{yao2018tip}. However, in this work, we mainly focus on deep learning based methods. 

One typical method, as illustrated in Fig.~\ref{fig1}~(a), is using ``loss correction'' to correct the loss of training samples based on the estimated noise transition matrix \cite{bootstrap,fcorrection}. The problem is that it is extremely difficult to get an accurate estimation of the noise transition matrix, thus inevitable false correction will lead to severe accumulated errors in the training process. Alternatively, as described in Fig.~\ref{fig1} (b), another popular training schema endeavors to adopt ``sample selection'' to identify and remove samples with label noise, and only use clean samples to update the networks \cite{decoupling,mentornet,coteaching}. Despite promising results that have been achieved in these approaches, the loss-based selection strategy favors ``easy'' examples. ``Hard'' and mislabeled samples (\eg, a ``Laysan Albatross'' image is labeled as ``Black Footed Albatross'') are ignored although they are surprisingly beneficial in making FG models more robust.

As shown in Fig.~\ref{fig1} (c), our idea is to re-utilize informative images (``hard'' and mislabeled examples) by selecting and correcting employable instances from high-loss samples. To be specific, we first split samples via the loss-based criterion. Low-loss instances are deemed to be clean, ``easy'' examples and their labels remain unaltered. Then, we perform a further partition on high-loss samples. In this separation process, we follow a simple but intuitive observation: for images of irrelevant categories, since they do not belong to any categories involved in the task, the network tends to be more confused when predicting their label probabilities. On the contrary, ``hard'' and mislabeled ones lean to obtain a relatively more certain prediction. After hard samples and mislabeled ones are selected out of high-loss instances, we manage to correct their labels and then feed them together with clean samples into the network for updating parameters. Extensive experiments and ablation studies on tasks of fine-grained image categorization demonstrate the superiority of our proposed approach over existing webly supervised state-of-the-art methods. The primary contributions of this work can be summarized as follows:

(1) A webly supervised deep model CRSSC is proposed to bridge the gap between FGVC tasks and numerous web images. Comprehensive experiments demonstrate that our approach outperforms existing state-of-the-art webly supervised methods by a large margin. 	

(2) Three types of FG web images (\textit{i.e.}, clean, reusable, and irrelevant) which inherently exist in collected web images are successfully identified and then separated by CRSSC. Compared with existing methods, our approach can further leverage reusable samples to boost the model learning.

(3) A novel label correction method which utilizes the prediction history of the network is proposed to re-label the reusable samples. Our experiments show that our label correction method is better than the existing one which uses the current epoch prediction results.

\section{Related Work}

\textbf{Fine-grained Visual Classification} Fine-grained visual classification (FGVC) aims to distinguish similar subcategories belonging to the same basic category. Generally, existing approaches can be roughly grouped into three categories: 1) strongly supervised methods, 2) weakly supervised methods, and 3) semi-supervised methods. Strongly supervised methods tend to require not only image-level labels but also manually annotated bounding boxes or part annotations \cite{branson2014,huang2016,wei2018}. Different from strongly supervised methods, weakly supervised methods cease to use bounding boxes and part annotations. Instead, methods in this group only require image-level labels during training \cite{lin2017,linmaji2017,gao2016,kong2017,cui2017,li2018,dubey2018,trilinear2019,destructionconstruction2019,complementarypart2019}. 
The third group involves leveraging web images in training the FGVC model \cite{cui2016fine,niu2018webly,xu2018webly}. However, these approaches still contain a certain level of human intervention, making them not purely web-supervised.

\noindent
\textbf{Webly Supervised Learning} Since learning directly from web images requires no human annotation, this learning scenario is becoming popular \cite{2016domain,2018discovering,zhang2020web,yao2018extracting,yao2019dynamically,2017new,2016automatic}. However, training deep FGVC models directly with web images usually leads to poor performance due to the existence of label noise and memorization effects \cite{zhang2016understanding,motivation2017} of neural networks. Existing deep methods for overcoming label noise can be categorized into two sets \cite{selfie2019}: 1) loss correction and 2) sample selection. Loss correction methods choose to correct the loss of training samples based on an estimated noise transition matrix \cite{bootstrap,fcorrection,goldberger2017,activebias,ren2018,pencil2019}. However, due to the difficulty in accurately estimating the noise transition matrix, accumulated errors induced by false correction are inevitable \cite{mentornet,coteaching}. Sample selection methods identify clean samples out of mini-batches based on their losses and leverage them to update the network \cite{decoupling,mentornet,coteaching,kumar2010,selfie2019}. Nevertheless, loss-based selection methods would cause the domination of easy samples in the training procedure while hard ones get ignored substantially \cite{activebias,harddatamining2016,focalloss2017,selfie2019}. Our proposed CRSSC can leverage additional reusable samples to boost the deep FGVC models.

\begin{figure*}[t]
\centering
\includegraphics[width=0.98\textwidth]{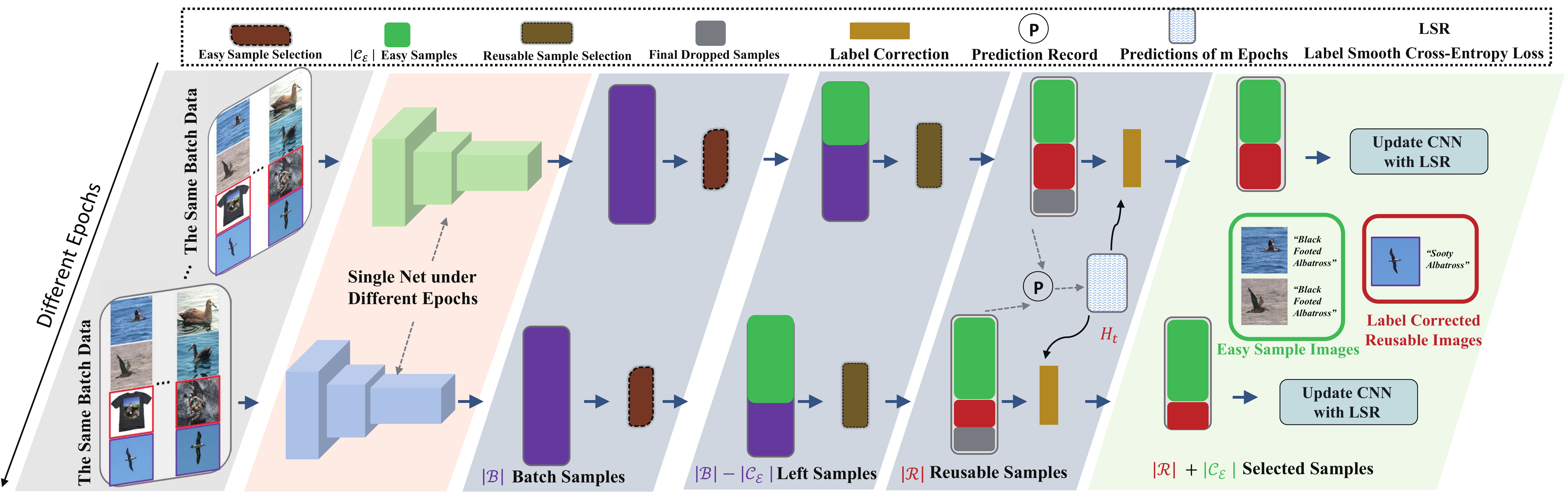}
\vspace{-0.2cm}
\caption{\small{The architecture of our proposed model. The deep network feeds forward a mini-batch of web images to predict labels and calculate losses for each image. Then the drop module sorts and selects small-loss instances as the easy sample set $\mathcal{C_E}$. Furthermore, we take a reuse module to leverage part of the high-loss instances. Specifically, the reuse module calculates the prediction certainty for each high-loss instance and selects high-certainty instances as reusable set $\mathcal{R}$. To use samples in $\mathcal{R}$, our model utilizes predictions of previous epochs to re-label reusable samples. Finally, training instances in $\mathcal{C_E} \cup \mathcal{R}$ are leveraged together to promote the optimization of model.}}
\vspace{-0.2cm}
\label{fig2}
\end{figure*}

\section{The Proposed Approach}

\subsection{Overview}

Fig.~\ref{fig2} presents the architecture of our proposed model. Generally, we can train a deep FGVC model through a well-labeled dataset $\mathcal{D} = \{(x_i, y_i) | 1 \le i \le M_L\}$, in which $x_i$ is the $i$-th training sample and $y_i$ is the corresponding ground-truth label. In the conventional training schema, the model parameters are updated by optimizing a cross-entropy loss $L_{CE}$ as follows:
\begin{equation}
\label{eq1}
\theta_{t+1} = \theta_t - \alpha\nabla [\frac{1}{|\mathcal{D}|} \sum_{x_i \in \mathcal{D}} L_{CE}(f(x_i;\theta_t) ,y_i)],
\end{equation}
where $\theta_t$ is the network parameter in the $t$-th training epoch and $f(x_i;\theta_t)$ is the network output of sample $x_i$.

However, for web images $\mathcal{D_W} = \{(x_i, \hat{y}_i) | 1 \le i \le M_W\}$, reliable labels are not always available and they are usually associated with label noise. Then we can divide $\mathcal{D_W}$ into three subsets:  
\begin{equation}
\label{eq2}
\mathcal{D_W} = \mathcal{N} \cup \mathcal{M} \cup \mathcal{C},
\end{equation}
where $\mathcal{N}$ indicates noisy samples, $\mathcal{M}$ represents mislabeled ones, and $\mathcal{C}$ stands for the clean set. More specifically, $\mathcal{C}$ can be further separated into easy example set $\mathcal{C_E}$ and hard example set $\mathcal{C_H}$:
\begin{equation}
\label{eq3}
\mathcal{C} = \mathcal{C_E} \cup \mathcal{C_H}.
\end{equation}
It should be noted that, as the training proceeds, hard samples will gradually become ``easy''. Thus, the split of the clean set $\mathcal{C}$ changes in the training process.
In this work, we aim to train a robust deep FGVC model through noisy web data. Our main idea is to properly select and then re-label informative training samples for boosting the robustness of the FGVC model. Based on the division described in Eq.~\eqref{eq2} and \eqref{eq3}, we regard the union of $\mathcal{C}$ and $\mathcal{M}$ as informative training set $\mathcal{I} = \mathcal{C} \cup \mathcal{M}$, though samples in $\mathcal{M}$ have to be corrected before being fed into the model for further network optimization.

\subsection{Drop and Reuse}

To optimize the network with only useful knowledge from the web images, we have to \textbf{1}) eliminate the negative influence from samples which belong to $\mathcal{N}$, \textbf{2}) reduce the misleading impact of samples belonging to $\mathcal{M}$. Therefore, two key challenges of tackling label noise in web images $\mathcal{D_W}$ are: \textbf{1}) how to select samples which belong to $\mathcal{I}$ and prevent the network from learning irrelevant samples, and \textbf{2}) how to correct labels of mislabeled ones and reuse them as part of the informative knowledge.

Memorization effects \cite{motivation2017,zhang2016understanding} indicate that, on noisy datasets, CNNs tend to first learn clean and easy patterns in initial epochs. As the number of epochs increases, CNNs will eventually overfit on noisy samples. Our key idea is to drop these noisy instances before they are memorized.
A widely used sample selection strategy is to separate instances based on losses \cite{mentornet,coteaching}. These methods typically select a human-defined proportion $ (1 - \tau) \times 100\% $ of low-loss instances as clean samples and directly drop the rest ones. Although significant improvements have been achieved in these works for dealing with label noise, the way they separate instances would lead to the mistaken deletion of samples belonging to $\mathcal{C_H}$ as hard examples also tend to produce high losses. Moreover, mislabeled samples would also be dropped by these sample selection methods due to their high losses.

To tackle drawbacks of these loss-based sample selection methods, we design a drop and reuse mechanism to further select useful instances from high-loss samples. Furthermore, for the purpose of avoiding involving the human-defined noise rate $ \tau $, we modify the conventional loss-based sample selection as in Definition~\ref{definition_drop}. Through adopting our proposed loss-based drop module as well as the reuse module, we can effectively avoid mistakenly deleting hard examples and can also make full use of mislabeled samples.

\begin{definition}
\label{definition_drop}
In a mini-batch $\mathcal{B}$, a sample $x \in \mathcal{B}$ belongs to $\mathcal{C_E}$ only if its loss $L_{CE}(x, \hat{y}; \theta) < \frac{1}{|\mathcal{B}|} \sum_{x_i \in \mathcal{B}} L_{CE}(x_i, \hat{y}_i; \theta)$
\end{definition}

We use the average loss as the selection threshold for dynamically separating informative samples from irrelevant ones. Specifically, due to limited robustness in initial epochs, more samples tend to have high losses. The selection threshold will be large and CNNs will learn easy patterns from as many samples as possible. As the training proceeds, CNNs gain more robust ability and more samples tend to have low losses. The selection threshold will be small and more samples will be dropped. In this situation, CNNs will discard as many samples as possible for ensuring the data learned by CNN is informative. In this way, our method can reduce the negative impact of error accumulation.

\begin{figure}[t]
\centering
\includegraphics[width=\linewidth]{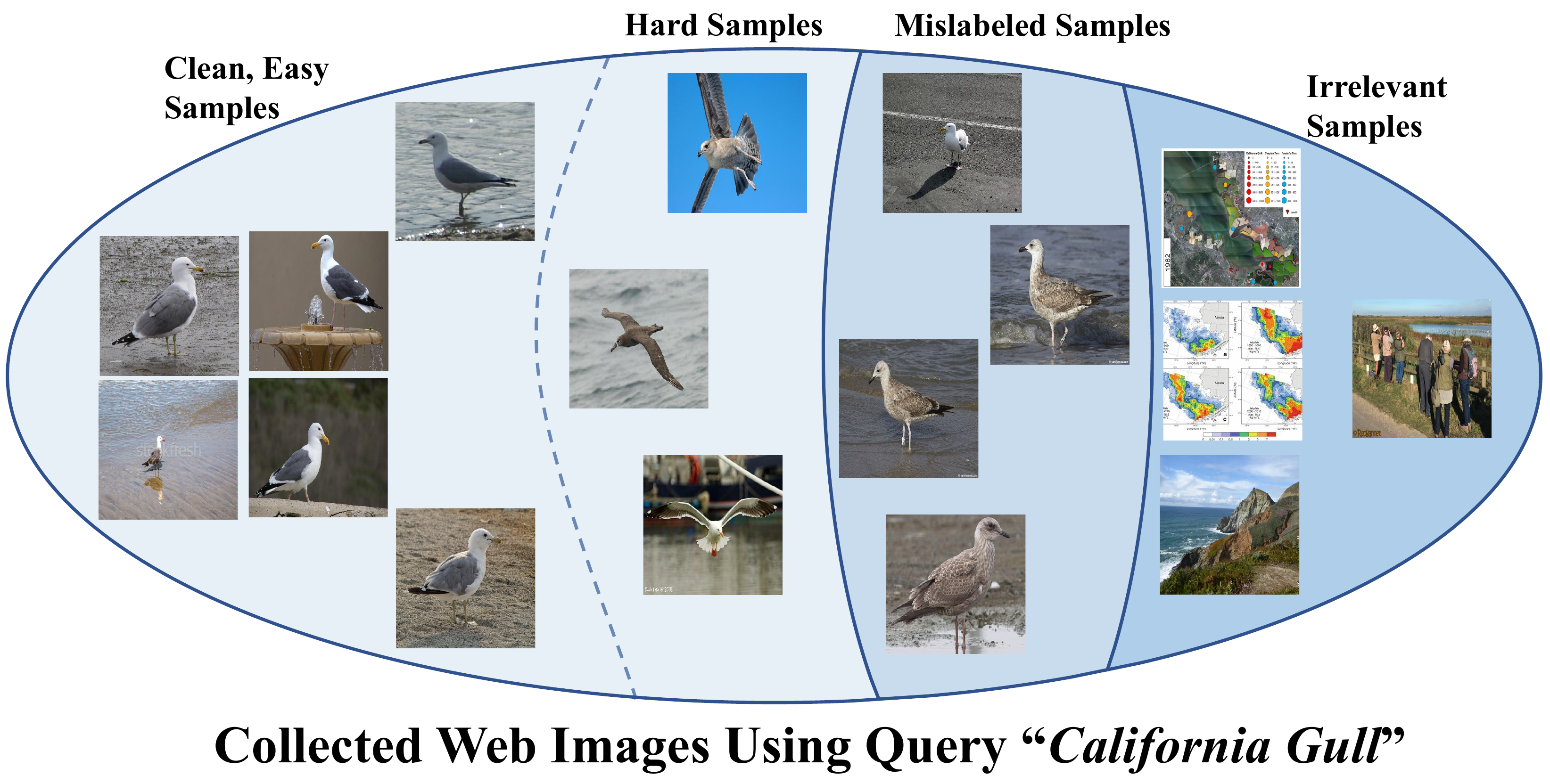}
\vspace{-0.4cm}
\caption{\small{The easy, hard, mislabeled, and irrelevant web images collected through query ``California Gull'', in which hard examples and mislabeled ones can be leveraged to boost the robustness of the deep FGVC model.}}
\vspace{-0.2cm}
\label{fig3}
\end{figure}

\textbf{Why can we distinguish reusable samples from irrelevant ones?} Intuitively, the predicted label probability of a reusable sample has a completely different pattern from that of an instance which belongs to an irrelevant category. Samples belonging to $\mathcal{N}$ can be distinguished based on the confusion of prediction. For example, as shown in Fig.~\ref{fig3}, in a bird classification task, when we feed a bird image (\eg, hard or mislabeled samples) into the model, the network tends to produce a certain prediction although it may have a high loss due to incorrect labeling. However, if we feed an irrelevant sample (\eg, a bird distribution map), which apparently belongs to $\mathcal{N}$ in this task, the network would get confused and thus produce an uncertain prediction. Inspired by this observation, we formalize a new criterion to further select reusable samples from the ones with high-loss: 

\begin{definition}
\label{definition_reuse}
In a mini-batch $\mathcal{B}$, a sample $x \in \mathcal{B} \cap \mathcal{C_E}^{-1}$ is reusable if its prediction certainty $V(x; \theta)$ satisfies the condition: $V(x; \theta) \ge \frac{1}{|\mathcal{B} \cap \mathcal{C_E}|} \sum_{x_i \in \mathcal{B} \cap \mathcal{C_E}} V(x_i; \theta)$.  
\end{definition}

\cite{chang2017active} demonstrated that the prediction variance can be used to measure the uncertainty of each sample in classification tasks. Therefore, in order to quantify the certainty of prediction, we simply adopt the standard deviation (the square root of the variance) of predicted probabilities for sample $x$:
\begin{equation}
\label{eq4}
V(x; \theta) = \sqrt{\frac{1}{K} \sum_{j=1}^{K}  (P_j(x; \theta) - \mu)^2},
\end{equation}
where $K$ is the number of categories, $P_j(x; \theta)$ is the softmax result of $f(x;\theta_t)$, which is deemed as the (pseudo) predicted probability of sample $x$ belonging to the $j$-th category, and $\mu$ is the mean value of predicted probabilities. Since $\sum_{j=1}^{K} P_j(x; \theta) = 1$, we can easily rewrite Eq.~\eqref{eq4} as:
\begin{equation}
\label{eq5}
V(x; \theta) = \sqrt{\frac{1}{K} \sum_{j=1}^{K} P_j(x; \theta)^2 - \frac{1}{K^2}}.
\end{equation}
The standard deviation of predicted probabilities is consistent with prediction certainty. From Eq.~\eqref{eq5} and the constraint of $\sum_{j=1}^{K} P_j(x; \theta) = 1$, we can have the following observations. \textbf{1}) The standard deviation of predicted probabilities $V(x; \theta)$ is bounded. \textbf{2}) $V(x; \theta)$ gets larger when one label's probability (\eg, $P_{q}(x, \theta)$) gets notably higher. \textbf{3}) If $P_{q}(x, \theta)$ is significantly higher than others' value, the network is more certain about its prediction on this sample. $V(x; \theta)$ reaches its maximum value $\frac{\sqrt{K-1}}{K}$ when 
\begin{equation}
\label{eq5_1}
\exists\: q \in \{1,2,3,...,K\}, P_{j}(x, \theta) = 
\left\{
\begin{array}{ll}
1, &j = q \\
0, &j \ne q
\end{array}.
\right.
\end{equation}
In this case, the prediction certainty also reaches its maximum. On the other hand, $V(x; \theta)$ gets smaller when all labels' probabilities get closer to each other. In this case, the prediction certainty also drops. This observation is consistent with the intuition. It results from a fact that if the network produces a prediction in which each label's probability is close to others, it means the network is highly confused about this sample.
$V(x; \theta)$ reaches its minimum value 0 when 
\begin{equation}
\label{eq5_2}
\forall j \in \{1,2,3,...,K\}, P_j(x, \theta) = 1/K.
\end{equation}
In this situation, the model fails as all labels are predicted equally.

\subsection{Label Correction}

The reusable samples selected by the certainty-based criterion defined in Definition~\ref{definition_reuse} are assembled into a sample set $\mathcal{R}$. To be specific, $\mathcal{R}$ includes two types of images: \textbf{1}) mislabeled instances and \textbf{2}) hard examples that have correct labels. 

In order to leverage these informative samples for training, their noisy labels have to be corrected. \cite{yang2018recognition} proposed to use the current prediction to replace original labels. Nevertheless, due to a lack of robustness for CNN predictions on noisy datasets, using a single prediction to relabel mislabeled instances may result in error accumulation. 
Therefore, different from \cite{yang2018recognition}, we propose to correct noisy labels using the prediction history. This results from a fact that averaging distributions over classifier iterations can increase stability and reduce the influence of misleading predictions. 
Empirical results show that our label correction method works better.

Specifically, we record the label prediction as well as its corresponding predicted probability for each training sample $x \in \mathcal{D_W}$. A history list $H_t(x)$ is defined as follows: 
\begin{equation}
\label{eq6}
H_t(x) = \{(h_{t-i}(x), p_{t-i}(x)) | 1 \le i \le m, x \in \mathcal{D_W}\},
\end{equation}
in which $h_t(x)$ is the label prediction of the sample $x$ in the $t$-th epoch, and $p_t(x)$ is its corresponding predicted probability. The history list $H_t(x)$ is maintained to memorize each sample's prediction of the previous $m$ epochs. 

When calculating forward losses, we correct samples belonging to $\mathcal{R}$ by replacing their original labels with corrected ones defined in Definition~\ref{definition_relabel}. For samples belonging to $\mathcal{C_E}$, we directly use their original labels. The leftover samples are regarded as irrelevant data and are excluded for robust training.

\begin{definition}
	\label{definition_relabel}
	For $x \in \mathcal{R}$, its corrected label $y^{corr}$ is the label prediction who has the highest accumulated probability in the previous $m$ epochs:
	\begin{equation}
	\label{eq7}
	y^{corr} = \underset{1 \le j \le K}{\arg\max} \sum_{i=1, h_{t-i}(x) = j}^m p_{t-i}(x).
	\end{equation}
\end{definition}

\textbf{Why don't we need to separate mislabeled and hard examples?} As stated above, both mislabeled samples and hard ones are included in $\mathcal{R} = \mathcal{C_H} \cup \mathcal{M}$. It is difficult to make a reliable split on $\mathcal{R}$ to distinguish hard samples and mislabeled ones. As a result, the label correction might also relabel hard samples using their previous predictions. However, both hard examples and mislabeled ones tend to produce consistent predictions on their true labels, though their predicted probabilities might be lower than those of easy samples. Therefore, using prediction history to relabel hard samples would not compromise the model.

\begin{algorithm}[t]\small
	\SetAlgoLined
	\KwInput{Initialized network $f$, warm-up epochs $T_w$, maximum epoch $T_{\max}$, and training set $\mathcal{D_W}$.}
	\For{$T = 1, 2, ..., T_{\max}$}
	{
		Randomly draw a mini-batch $\mathcal{B}$ from $\mathcal{D_W}$. \\
		\eIf {$T_w \le T \le T_{\max}$}{ 
			Construct $\mathcal{B_{C_E}} = \mathcal{B} \cap \mathcal{C_E}$ through Definition~\ref{definition_drop}. \\
			Construct $\mathcal{B_R} = \mathcal{B} \cap \mathcal{R}$ by Definition~\ref{definition_reuse}. \\
			Relabel samples in $\mathcal{B_R}$ with Definition~\ref{definition_relabel}. \\
			Update the network $f$ with Eq.~\eqref{eq8}.
		}
		{Update the network $f$ according to Eq.~\eqref{eq1}. \\
		}	
	}
	\KwOutput{Updated network $f$.} 
	\caption{The proposed CRSSC algorithm}
	\label{alg}
\end{algorithm}

\subsection{Summary of CRSSC}

Based on the Definition~\ref{definition_reuse} and \ref{definition_relabel}, we can reformulate the network update function in Eq.~\eqref{eq1} as 
\begin{equation}
\label{eq8}
\theta_{t+1} = \theta_t - \alpha\nabla [\frac{1}{|\mathcal{C_E} \cup \mathcal{R}|} \sum_{x_i \in \mathcal{C_E} \cup \mathcal{R}} L_{CE}(f(x_i;\theta_t) ,y_i)],
\end{equation}
where 
\begin{equation}
\label{eq9}
y^*_i = \left\{
\begin{aligned}
& \hat{y}_i  ,& x_i \in \mathcal{C_E}; \\
& y^{corr}_i ,& x_i \in \mathcal{R}.
\end{aligned}
\right.
\end{equation}
Here, to further enhance the generalization performance of CRSSC, we adopt the Label Smoothing Regularization \cite{szegedy2016} when calculating the cross-entropy loss. That is to say, for input image $x_i$, we adopt the following smoothed ground-truth probability $\{q_j | 1 \le j \le K\}$ in the loss calculation:
\begin{equation}\label{eq10}
q_j = 
\left\{
\begin{array}{ll}
1- \epsilon,         & {j = y^*_i}    \\
\epsilon / (K-1),    & {j \neq y^*_i}
\end{array}.
\right.
\end{equation}

For CRSSC, we first train the network on whole training set in a conventional manner to warm up the network. The reason is that deep CNN has memorization effects \cite{zhang2016understanding,motivation2017} and will learn clean and easy patterns in the initial epochs. With this warm-up step, CNN will get equipped with an initial learning capacity. Then, we perform a two-step sample selection in each mini-batch $\mathcal{B}$: \textbf{1}) select low-loss instances from $\mathcal{B}$ using the criterion defined in the Definition~\ref{definition_drop}, and \textbf{2}) identify reusable samples, which have high prediction certainty in high-loss instances based on the Definition~\ref{definition_reuse}. Subsequently, reusable samples are relabeled based on the Definition~\ref{definition_relabel}. Finally, parameters of the network are updated using the clean, easy sample set $\mathcal{B_{C_E}}$ along with the reusable sample set $\mathcal{B_R}$. 
The detailed process of our proposed CRSSC is shown in Algorithm~\ref{alg}.

\section{Experiments}

\begin{table*}[t]\small
	\renewcommand{\arraystretch}{1.1}
	\centering
	\caption{\small{The ACA (\%) results on three benchmark datasets. BBox/Part means bounding box or part annotation is required during training. Training set means the training data is manually labeled (anno.) or collected from the web (web). iNat refers to the iNat2017 dataset.}}
	\vspace{-0.2cm}
	\begin{tabular}{c|r|c|c|c|c|c}
		\hline
		\multirow{2}{*}{\textbf{Types}} & \multirow{2}{*}{\textbf{Methods}\:\:\:\:\:\:\:} & \multirow{2}{*}{\textbf{BBox/Part}} & \multirow{2}{*}{\textbf{Training Set}} & \multicolumn{3}{c}{\textbf{Performance}} \\
		\cline{5-7}  &    &     &     & FGVC-Aircraft \cite{fgvc-aircraft}  &  CUB200-2011 \cite{cub200-2011}  & Stanford Cars \cite{stanford-cars}\\
		\hline
		\multirow{4}{*}{Strongly}		
		& Part-Stacked \cite{huang2016}     		&    \checkmark    &  anno.   &  -        &  76.6            &  -     \\
		& Coarse-to-fine \cite{coarse-to-fine}  &    \checkmark    &  anno.   & 87.7      &  82.9            & -      \\
		& HSnet   \cite{hsnet2017}              &    \checkmark    &  anno.   &  -        &  87.5            &  93.9  \\
		& Mask-CNN  \cite{wei2018}              &    \checkmark    &  anno.   &  -        &  85.7            &  -     \\
		\hline
		\multirow{4}{*}{Weakly}
		& iSQRT-COV \cite{li2018}               &                  &  anno.   &  91.4     &  88.7            &  93.3  \\
		& Parts Model \cite{complementarypart2019}&                &  anno.   &  -        &  90.4            &  -     \\
		& TASN \cite{trilinear2019}             &                  &  anno.   &  -        &  89.1            &  93.8  \\
		& DCL \cite{destructionconstruction2019}&                  &  anno.   &  93.0     &  87.8            &  94.5  \\
		\hline
		\multirow{4}{*}{Semi}
		& Cui \al \cite{cui2016fine}          &                  & anno.+web&  -        &  89.7            &  -     \\
		& Xu \al \cite{xu2018webly}           &                  & anno.+web&  -        &  84.6            &  -     \\
		& Niu \al \cite{niu2018webly}         &                  & anno.+web&  -        &  76.5            &  -     \\
		& Cui \al \cite{cui2018large}         &                  & anno.+iNat&  90.7     &  89.3            &  93.5  \\
		\hline
		\multirow{6}{*}{Webly}
		& VGG-16 \cite{vgg}                     &                  &  web     &  68.4     & 66.3            &  61.6   \\
		& ResNet-50 \cite{resnet}		        &                  &  web     &  60.4     & 64.4            &  60.6   \\
		& B-CNN \cite{lin2017}                  &                  &  web     &  64.3     & 66.6            &  67.4   \\
		& Decoupling \cite{decoupling}          &                  &  web     &  75.9      & 70.6          &  75.0   \\
		& Co-teaching \cite{coteaching}         &                  &  web     &  72.8     &  73.9           &  73.1   \\
		& \textbf{CRSSC\quad\:}                 &                  &  web     &  \textbf{76.5}     & \textbf{77.4}   &  \textbf{76.6}   \\
		\hline	
	\end{tabular}
	\label{tab1}
\end{table*}

\subsection{Datasets and Evaluation Metrics}

\textbf{Datasets:} We evaluate our approach on three popular benchmark fine-grained datasets: CUB200-2011 \cite{cub200-2011}, FGVC-Aircraft \cite{fgvc-aircraft} and Stanford Cars \cite{stanford-cars}. \\
\textbf{Evaluation Metric:} Average Classification Accuracy (\textbf{ACA}) is taken as the default evaluation metric.

\subsection{Implementation Details}

\begin{table*}[t]\small
	\begin{minipage}{0.48\linewidth}
		\centering
		\renewcommand\tabcolsep{7pt}
		\renewcommand{\arraystretch}{1.1}
		\caption{\small{The ACA ($\%$) comparison by using different backbones.}}
		\vspace{-0.2cm}
		\begin{tabular}{c|c|c|c|c}
			\hline			
			\textbf{ Backbone }   &   \textbf{ Method }		&  \textbf{ Aircraft }      &  \textbf{ CUB200 }     &    \textbf{ Cars }     \\
			\hline
			\multirow{2}{*}{VGG-16} 
			&		Standard             &  68.4              &  66.3                 &  61.6        \\
			&		CRSSC                &  78.3              &  77.8                 &  81.6        \\
			\hline
			\multirow{2}{*}{ResNet-18} 
			&		Standard             &  53.7              &  59.3                 &  55.6        \\
			&		CRSSC                &  75.6              &  76.8                 &  84.2        \\
			\hline
			\multirow{2}{*}{ResNet-50} 
			&		Standard             &  60.8              &  64.4                 &  60.6        \\
			&		CRSSC                &  82.5              &  81.3                 &  87.7        \\  
			\hline	
		\end{tabular}
		\label{tab:backbone}
	\end{minipage}
	\hspace{0.033\linewidth}
	\begin{minipage}{0.48\linewidth}
		\centering
		\renewcommand\tabcolsep{7pt}
		\renewcommand{\arraystretch}{1.1}
		\caption{\small{The ACA ($\%$) comparison by using different steps.}} 
		\vspace{-0.2cm}
		\begin{tabular}{l|c}
			\hline
			\textbf{\:Model}                                                                                                    &   \textbf{ACA ($\%$)}            \\
			\hline
			\:ResNet-18                                                                                   &   59.3             \\
			\:+ Def.~\ref{definition_drop}                                                                                      &   69.7             \\  
			\:+ Def.~\ref{definition_drop} + Def.~\ref{definition_reuse}                                                        &   69.9             \\          
			\:+ Def.~\ref{definition_drop} + Def.~\ref{definition_reuse} + Def.~\ref{definition_relabel}                        &   73.4             \\      
			\:+ Def.~\ref{definition_drop} + Def.~\ref{definition_reuse} + Def.~\ref{definition_relabel} + LSR (CRSSC)          &   75.6             \\  
			\:+ Fine-tuned CRSSC                                                                                                &   76.8             \\  
			\hline
		\end{tabular}
		\label{tab:module_stacking}
	\end{minipage}
\end{table*}

\begin{figure*}[t]
	\begin{minipage}{0.47\linewidth}
		\centering
		\includegraphics[width=1\textwidth]{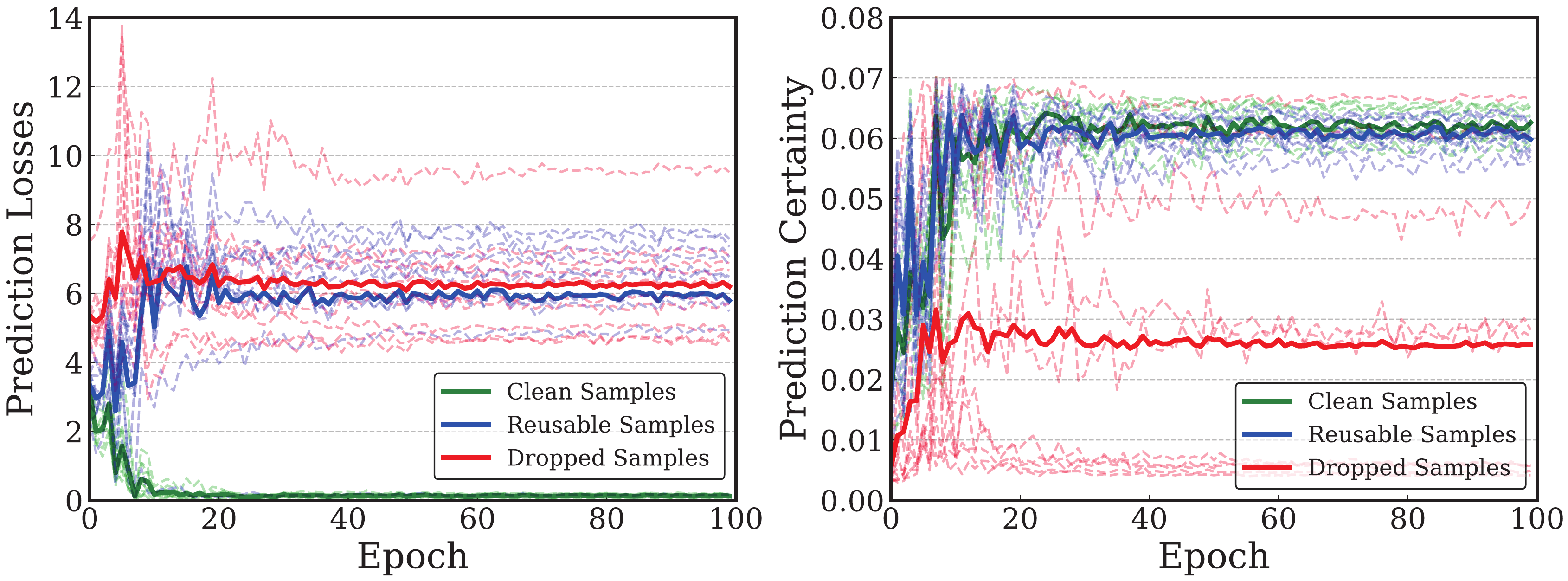}
		\vspace{-0.5cm}
		\caption{\small{The prediction loss (\textbf{left}) and certainty (\textbf{right}) of clean, reusable, and dropped samples as training progresses. The value on the individual image is plotted in the dotted line and the average value of each group is plotted in the solid line.}}
		\label{fig4}
		\vspace{-0.2cm}
	\end{minipage}
	\hspace{0.04\linewidth}
	\begin{minipage}{0.47\linewidth} 
		\centering
		\includegraphics[width=1\textwidth]{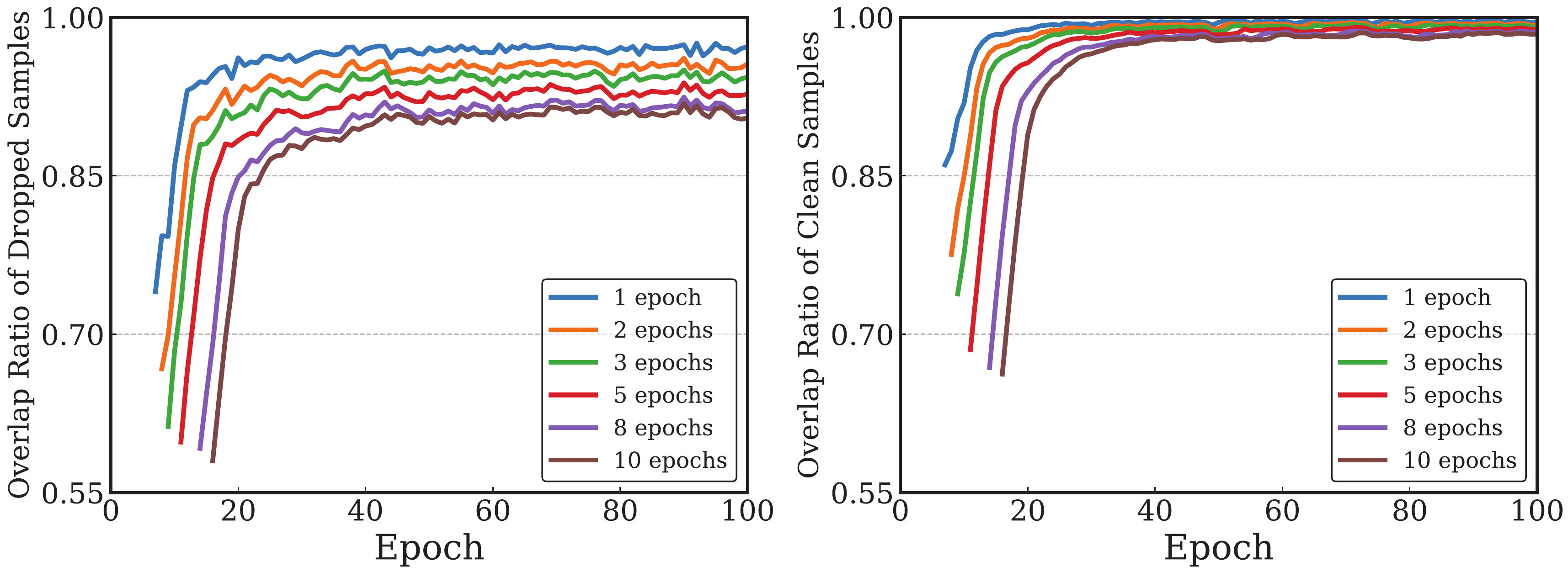}
		\vspace{-0.5cm}
		\caption{\small{The sample selection overlap of dropped samples (\textbf{left}) and clean ones (\textbf{right}) between each epoch and its \textbf{previous} 1, 2, 3, 5, 8, 10 epochs. The comparison starts after the warm-up epochs, \ie, the blue line starts from the 7th epoch.}}
		\label{fig5}
		\vspace{-0.2cm}
	\end{minipage}
\end{figure*}

\textbf{Data preparation:} To collect the web training set, we follow \cite{niu2018webly} and retrieve images from image search engine using the category labels in benchmark datasets. 
For ensuring no overlaps between the training and testing set, we additionally perform a PCA near-duplicate removal \cite{pcaduplicateremoval2016} between collected web images and test images in the benchmark datasets. Finally, we regard filtered web images (13503 for FGVC-Aircraft, 18388 for CUB200-2011, and 21448 for Stanford Cars) as the training set and adopt testing images from original benchmark datasets.\\
\textbf{CRSSC learning:} We use a pre-trained model (\eg, VGG-16 \cite{vgg}) as the basic CNN network. 
The number of warm-up epochs $T_w$ is tested in $\{5, 8, 10, 15, 20\}$. 
The history list length limit $l$ is selected from $\{0, 5, 10, 15, 20\}$. 
The LSR parameter $\epsilon$ is chosen from $\{0.0,0.2,0.5,0.6,0.8\}$. 
During the network optimization, we adopt a SGD optimizer with momentum $= 0.9$. The learning rate, batch size, and weight decay are set to be 0.01, 32, and 0.0003, respectively.

\subsection{Baseline Methods}

\textbf{Strongly supervised methods} require bounding boxes or part annotations during training. This set of baselines includes Part-Stacked \cite{huang2016}, Coarse-to-fine \cite{coarse-to-fine}, HSnet \cite{hsnet2017}, and Mask-CNN \cite{wei2018}. \textbf{Weakly supervised methods} require image-level labels, including Parts Model \cite{complementarypart2019}, iSQRT-COV \cite{li2018}, TASN \cite{trilinear2019}, and DCL \cite{destructionconstruction2019}. \textbf{Semi-supervised methods} leverage web images but remain involving human intervention, including Cui \al \cite{cui2016fine}, Xu \al \cite{xu2018webly}, Niu \al \cite{niu2018webly}, and Cui \al \cite{cui2018large}. For strongly, weakly, and semi-supervised methods, we report the performances in their papers.\\
\textbf{Webly supervised methods} directly leverage web images without human involvement, including VGG-16 \cite{vgg}, ResNet-50 \cite{resnet}, B-CNN \cite{lin2017}, Decoupling \cite{decoupling} and Co-teaching \cite{coteaching}. 
To be fair, we use the same backbone B-CNN \cite{lin2017} in Decoupling, Co-teaching and our CRSSC. For basic networks VGG-16, ResNet-50, and B-CNN, we fine-tune them with noisy web images. 

\subsection{Experimental Results}

Table~\ref{tab1} presents the comparison of ACA results on three benchmark datasets. It should be noted that the results of webly methods are all produced from experiments using exactly the same training data.
By observing Table~\ref{tab1}, we can notice that our approach performs better than other webly supervised methods on all three benchmark datasets. Compared with basic networks VGG-16, ResNet-50, and B-CNN, our CRSSC (with backbone B-CNN) can effectively alleviate the influence of label noise in the process of model training. Compared with state-of-the-art webly supervised methods Decoupling and Co-teaching, our approach can additionally identify reusable samples and salvage them by performing a label correction. Thus, our CRSSC can efficiently explore more useful samples to boost the robustness of the FGVC model.

\section{Ablation Studies}

\subsection{Training Loss and Prediction Certainty}

The prediction loss and certainty are two fundamental criteria for selecting informative samples. To investigate the distribution of prediction loss and certainty for clean, reusable, and dropped samples in training process, we select 30 instances in total, 10 images for each group (clean, reused, and dropped) and plot their prediction losses as well as prediction certainties. The experimental results are shown in Fig.~\ref{fig4}. 

By observing Fig.~\ref{fig4}, we can find that as the network training forwards, the losses of clean samples decrease sharply while their prediction certainties increase steadily. Regarding reusable samples, although some of them have a fairly higher loss than clean ones, their prediction certainties increase remarkably as training progresses. The explanation is that reusable samples are either hard or mislabeled instances, thus tend to produce confident predictions consistently.
The high loss and low prediction certainty of dropped samples demonstrate that our CRSSC can successfully identify and drop these irrelevant samples.

\begin{table*}[t]\small
	\begin{minipage}{0.48\linewidth}
		\renewcommand\tabcolsep{7pt}
		\renewcommand{\arraystretch}{1.1}
		\centering
		\caption{\small{The ACA ($\%$) results of combining CRSSC with Co-teaching in same and different backbones on CUB200-2011.}}
		\vspace{-0.2cm}
		\begin{tabular}{c|c|c|c|c}
			\hline
			\textbf{Dataset} &     \textbf{Net 1}       &  \textbf{ACA 1}     &     \textbf{Net 2}       &    	\textbf{ACA 2}   \\
			\hline
			\multirow{3}{*}{CUB200-2011}         
			&     ResNet-18    &  77.4      &     ResNet-18    &    78.7                 \\
			&     ResNet-18    &  78.8      &     ResNet-50    &    81.9                 \\
			&     ResNet-50    &  81.9      &     ResNet-50    &    81.0                 \\
			\hline		
		\end{tabular}
		\label{tab:hybrid_coteaching}
	\end{minipage}
	\hspace{0.032\linewidth}
	\begin{minipage}{0.48\linewidth}
		\centering
		\renewcommand\tabcolsep{7pt}
		\renewcommand{\arraystretch}{1.1}
		\caption{\small{The ACA ($\%$) results of our CRSSC training on the combined datasets (web+labeled).}}
		\vspace{-0.2cm}
		\begin{tabular}{c|c|c|c}
			\hline			
			\textbf{ Backbone }	&  \textbf{FGVC Aircraft}       &  \textbf{CUB200}           &  \textbf{Stanford Cars}      \\
			\hline                       
			VGG-16         &  88.4           &  85.7           &  92.4   \\
			ResNet-18      &  88.4           &  86.8           &  92.4     \\
			ResNet-50      &  93.4           &  87.7           &  94.0     \\  
			\hline	
		\end{tabular}
		\label{tab:semi_supervision}
	\end{minipage}
\end{table*}

\begin{figure}[t]
\centering
\includegraphics[width=\linewidth]{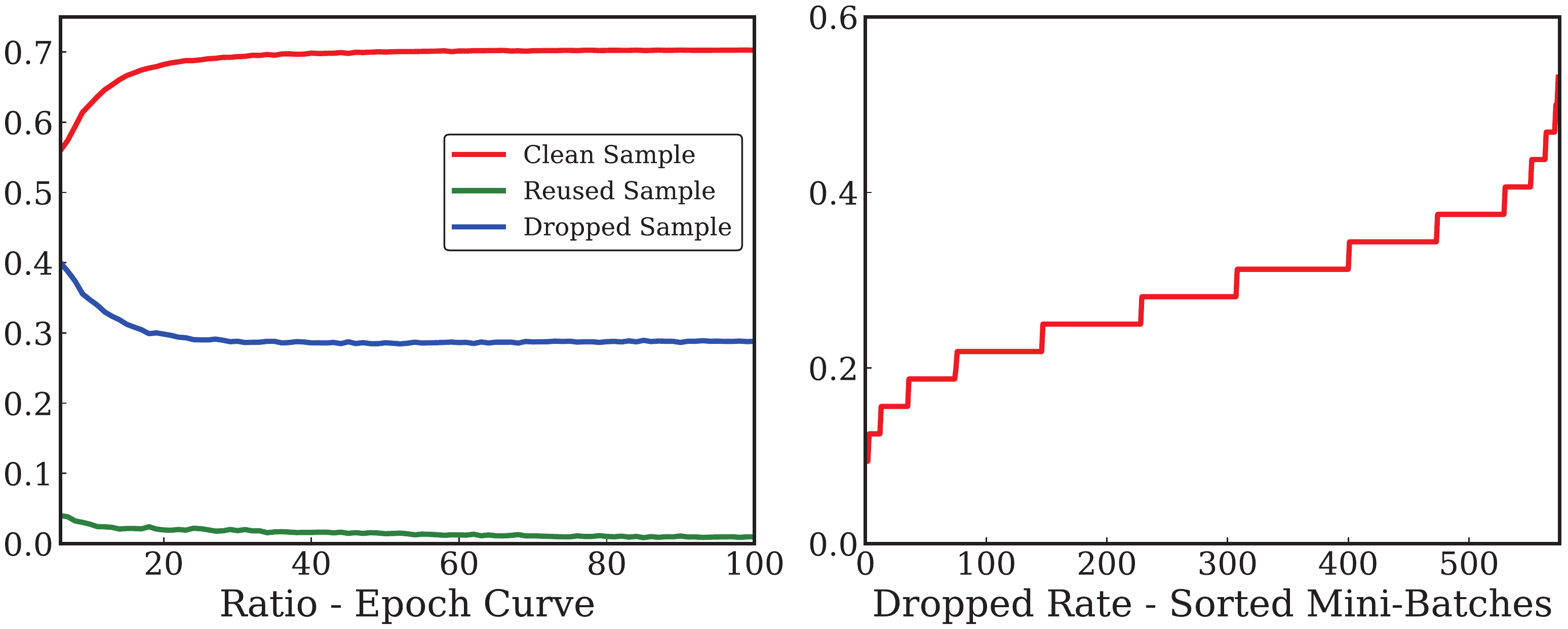}
\vspace{-0.5cm}
\caption{\small{The variations of clean, reusable, and dropped samples during the training processes (\textbf{left})}. The drop rate among sorted mini-batches in one epoch (\textbf{right}).}
\label{fig6}
\vspace{-0.45cm}
\end{figure}

\subsection{Overlap of Identified Samples in Epochs}

To further investigate the robustness of our sample selection strategy, we explore the overlap ratio of selected samples between adjacent epochs. To this end, we record the sample selection overlap between each epoch and its previous 1, 2, 3, 5, 8, 10 epochs. 
Let $\mathcal{C}^i$ represent the selected dropped (clean) sample set in the $i$-th epoch, Fig.~\ref{fig5} presents the sample selection overlap of dropped (clean) samples among the current $i$-th epoch and its previous $t$ epochs $\mathcal{F}_{t}^i = \mathcal{C}^i \cap \mathcal{C}^{i-1} \cap ... \cap \mathcal{C}^{i-t+1} \cap \mathcal{C}^{i-t}$. From Fig.~\ref{fig5}, we can observe that both dropped and clean samples grow steadily and finally converge to a high level, which firmly proves the consistency and robustness of our sample selection strategy. 

\subsection{Influence of Different Backbones}
It is well known that the choice of CNN architectures has a critical impact on object recognition performance. To investigate the influence of different backbones, we conduct experiments by using different basic networks VGG-16 \cite{vgg}, ResNet-18, and ResNet-50 \cite{resnet}. The experimental results are shown in Table~\ref{tab:backbone}. 

From Table~\ref{tab:backbone}, we can have the following observations: \textbf{1}) with a much deeper backbone network like ResNet-50, our CRSSC can yield significantly better performance than ResNet-18 and VGG-16. \textbf{2}) When training a basic network directly with noisy web images, the basic network with higher capacity may produce a worse result. However, by adopting our CRSSC, we can make full use of the learning capacity of basic networks via properly selecting reusable samples and correcting their labels. Compared with the standard network, the improvement of performance demonstrates the superiority of our proposed approach.

\subsection{Influence of Different Steps}

In this subsection, we investigate the influence of various steps on a basic network like ResNet-18. We first add the Def.~\ref{definition_drop} on the ResNet-18 network to construct a baseline. We then add the Def.~\ref{definition_drop} and \ref{definition_reuse} to construct another baseline. For the third baseline, we add all the Def.~\ref{definition_drop}, \ref{definition_reuse}, and \ref{definition_relabel} to the ResNet-18. For the fourth baseline, we add the label smoothing technique to complete our CRSSC method. Finally, we present a fine-tuned CRSSC model as the last baseline. The experimental results on the CUB200-2011 dataset are summarized in Table \ref{tab:module_stacking}. By observing Table \ref{tab:module_stacking}, we can find that the fine-tuned CRSSC framework obtains the best performance.

\begin{figure}
	\centering
	\includegraphics[width=\linewidth]{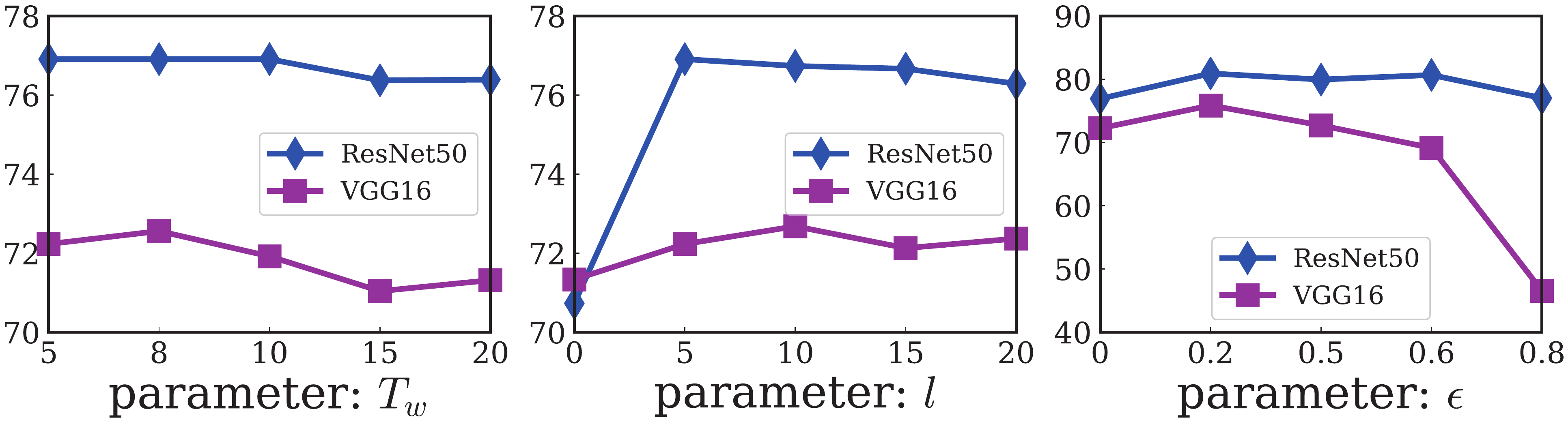}
	\vspace{-0.5cm}
	\caption{\small{The parameter sensitivities of warm-up epochs $T_w$ (\textbf{\textbf{left}}), history list length $l$ (\textbf{\textbf{middle}}), and the LSR smoothing level $\epsilon$ (\textbf{right}) by using different backbones VGG-16 and ResNet-50.}}
	\label{fig7}
	\vspace{-0.2cm}
\end{figure}

\subsection{Combining with Co-teaching}

Our proposed method is flexible with regard to combining with other techniques since it only involves sample selection and relabeling. Here, we combine our CRSSC with a state-of-the-art method Co-teaching \cite{coteaching} for further performance improvement. Following the setup in Co-teaching, we maintain two networks simultaneously. In each mini-batch, each network constructs its own $\mathcal{B_{C_E}}$ and $\mathcal{B_R}$ and subsequently feeds them into its peer network for further updating. Different from Co-teaching which only trains model with clean, easy samples, the combined method additionally leverages hard and mislabeled samples to promote the network optimization. Table~\ref{tab:hybrid_coteaching} demonstrates the ACA results of combining CRSSC with Co-teaching in same and different backbones on CUB200-2011 dataset. Compared with the naive CRSSC (presented in Table~\ref{tab:backbone}), we can observe that great improvement has been achieved in Table~\ref{tab:hybrid_coteaching}. 

\subsection{Combining Web and Labeled Data}

One of the roadblocks that limit the performance of fine-grained visual classification is the lack of enough labeled training data. The widely-used FGVC benchmark datasets (\eg, FGVC-Aircraft, CUB200-2011, and Stanford Cars) all suffer from limited training data, which severely prevented the FGVC task from being sufficiently benefited from the high learning capability of deep CNN. Therefore, employing web images as a supplement to existing fine-grained datasets also attracts considerable attention in recent years. Following the semi-supervised manner, we leverage collected web images as data augmentation to the labeled training data for training deep FGVC model. The experimental results of our CRSSC training on the combined data are shown in Table~\ref{tab:semi_supervision}.

\begin{figure}[t]
\centering
\includegraphics[width=\linewidth]{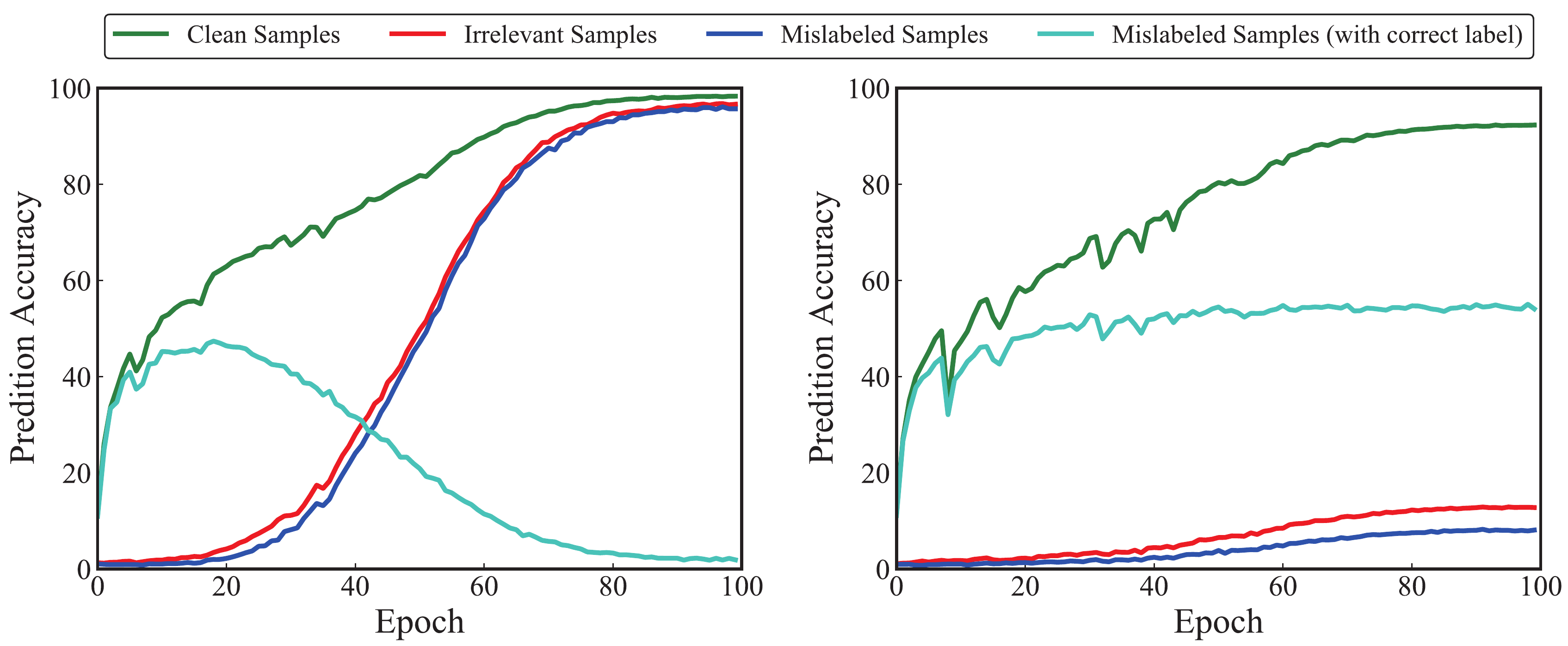}
\vspace{-0.5cm}
\caption{\small{The prediction accuracy ($\%$) of the baseline model (left) and our CRSSC model (right) during the process of training. The ``Mislabeled Samples'' curve (blue) represents the prediction accuracy with regard to corrupted labels while the ``Mislabeled Samples (with correct label)'' curve (cyan) denotes the prediction accuracy with regard to ground-truth labels.}}
\vspace{-0.2cm}
\label{fig8}
\end{figure}

\subsection{Trend of Samples in Training}

We present the ratio variations of identified clean, reusable, and dropped samples during the training processes in Fig.~\ref{fig6} (left). From Fig.~\ref{fig6} (left), we can notice that the ratio of clean samples increases steadily until convergence while training progresses. As the training continues, the previous hard examples will gradually become ``easy'' for our model. On the contrary, as the network training proceeds, the certainty-based criterion will gradually reduce the mistaken dropping, thus leading to a firm decrease in the ratio of dropped samples until convergences to the ground-truth noise rate. 
Additionally, since hard examples get fewer with the growth of network capability, the ratio of reusable samples also decreases, until only mislabeled examples are left. The final convergence of three groups demonstrates the stability and robustness of our sample selection strategy. Fig.~\ref{fig6} (right) shows the drop rate among sorted mini-batches in one epoch. From Fig.~\ref{fig6} (right), we can find the imbalance of dropped sample ratios across each mini-batch, which proves the necessity of avoiding using a predefined drop rate.

\subsection{Parameters in Proposed Approach}

For the parameters analysis, we concern three parameters, including \textbf{1}) the number of warm-up epochs $T_w$, \textbf{2}) the length of history list $l$, and \textbf{3}) the LSR smoothing level $\epsilon$. Fig.~\ref{fig7} gives the results on CUB200-2011 dataset. 

From Fig.~\ref{fig7} (left), we can observe that CRSSC is relatively stable when varying $T_w$ by fixing other two parameters. Both cases achieve the best performance when $T_w$ is selected in $[5, 8]$ and we select $T_w=5$ as the default option. The length of history list affects the precision of label correction. Intuitively, a higher value of $l$ may benefit the label correction. However, the relabeling could also be misled due to poor predictions of early epochs. From Fig.~\ref{fig7} (middle), we notice that the best performance can be obtained when $l \in [5, 10]$, we select $l=5$ as the default option. It should be noted that when $l$ = 0, this means that we use the prediction label of the current epoch instead of that with the highest accumulated probability in the previous $m$ epochs.

The smoothing level has an influence on the generalization ability. Compared with the case in which LSR is not leveraged (\textit{i.e.}~$\epsilon = 0$), from Fig.\ref{fig7} (right), we can find that the classification accuracy increases considerably when adopting a proper level of label smoothing. When $\epsilon \in [0.2, 0.6)$, the performance is fairly robust. However, as the $\epsilon$ gets larger than 0.6, the performance starts to decrease. This probably results from the fact that too large $\epsilon$ leads to a lack of proper ground-truth guidance in the training process.

\begin{figure}[t]
	\centering
	\includegraphics[width=\linewidth]{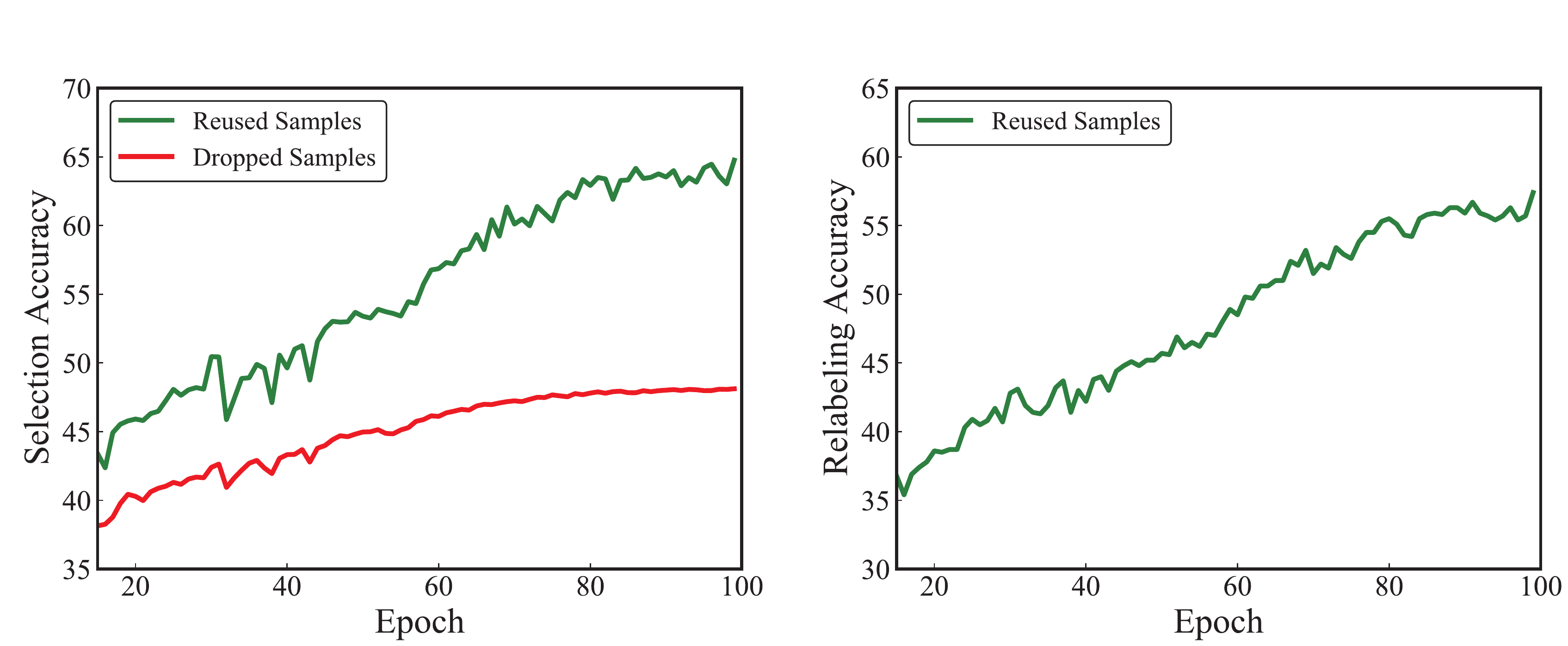}
	\vspace{-0.5cm}
	\caption{\small{The samples selection accuracy ($\%$) (left) and the sample relabeling accuracy ($\%$) (right) during the training process.}}
	\vspace{-0.2cm}
	\label{fig9}
\end{figure}

\subsection{Further Studies on Noisy-CIFAR100}

To further explore the effectiveness of our approach, we follow Co-teaching \cite{coteaching} and generate a synthetic dataset based on CIFAR100 \cite{krizhevsky2009} for further study. We first regard the last 20 categories of CIFAR100 as the irrelevant categories. Then we randomly select $20\%$ of the remaining training samples and corrupt their labels to simulate mislabeled data. We named this synthetic dataset as Noisy-CIFAR100. \\
\textbf{Prediction Accuracy for Different Samples:} As the memorization effects \cite{motivation2017,zhang2016understanding} indicated, CNN tends to fit clean samples in initial epochs and will eventually fit noise data (\ie, irrelevant samples, and mislabeled ones). 
Fig.~\ref{fig8} presents the prediction accuracy of baseline model ResNet-18 (left) and our CRSSC (right) model. By observing Fig.~\ref{fig8} (left), we can observe that, while the prediction accuracy on clean training samples grows steadily in the training process, CNN eventually fits noisy training data and degrades the classification ability of the final model. From Fig.~\ref{fig8} (right), we can notice that, by using our CRSSC, the over-fitting to noise data are effectively suppressed. Besides, by using our relabeling strategy, mislabeled samples can be better learned and finally contribute to boosting the model classification ability.\\
\textbf{Samples Selection and Relabeling Accuracy:} Fig.~\ref{fig9} (left) shows the samples selection accuracy of our approach and (right) presents the samples relabeling accuracy. From Fig.~\ref{fig9} (left), we can find that the samples selection accuracy (including both reused and dropped samples) grows steadily as the training proceeds. In addition, the samples relabeling accuracy also has a steady increases in training. 

\section{Conclusion}

In this work, we studied the problem of training FGVC models directly with noisy web images. Accordingly, we proposed a simple yet effective approach, termed as CRSSC, which trained a deep neural network using additionally selected hard and mislabeled samples to boost the robustness of the model. Comprehensive experiments showed that our approach has achieved state-of-the-art performance, compared with existing webly supervised methods.

\section*{Acknowledgments}
This work was supported by the National Natural Science Foundation of China (No. 61976116) and Fundamental Research Funds for the Central Universities (No. 30920021135).



\begin{thebibliography}{64}


\ifx \showCODEN    \undefined \def \showCODEN     #1{\unskip}     \fi
\ifx \showDOI      \undefined \def \showDOI       #1{#1}\fi
\ifx \showISBNx    \undefined \def \showISBNx     #1{\unskip}     \fi
\ifx \showISBNxiii \undefined \def \showISBNxiii  #1{\unskip}     \fi
\ifx \showISSN     \undefined \def \showISSN      #1{\unskip}     \fi
\ifx \showLCCN     \undefined \def \showLCCN      #1{\unskip}     \fi
\ifx \shownote     \undefined \def \shownote      #1{#1}          \fi
\ifx \showarticletitle \undefined \def \showarticletitle #1{#1}   \fi
\ifx \showURL      \undefined \def \showURL       {\relax}        \fi
\providecommand\bibfield[2]{#2}
\providecommand\bibinfo[2]{#2}
\providecommand\natexlab[1]{#1}
\providecommand\showeprint[2][]{arXiv:#2}

\bibitem[\protect\citeauthoryear{Arpit, Jastrz{\k{e}}bski, Ballas, Krueger,
  Bengio, Kanwal, Maharaj, Fischer, Courville, Bengio, et~al\mbox{.}}{Arpit
  et~al\mbox{.}}{2017}]%
        {motivation2017}
\bibfield{author}{\bibinfo{person}{Devansh Arpit},
  \bibinfo{person}{Stanis{\l}aw Jastrz{\k{e}}bski}, \bibinfo{person}{Nicolas
  Ballas}, \bibinfo{person}{David Krueger}, \bibinfo{person}{Emmanuel Bengio},
  \bibinfo{person}{Maxinder~S Kanwal}, \bibinfo{person}{Tegan Maharaj},
  \bibinfo{person}{Asja Fischer}, \bibinfo{person}{Aaron Courville},
  \bibinfo{person}{Yoshua Bengio}, {et~al\mbox{.}}}
  \bibinfo{year}{2017}\natexlab{}.
\newblock \showarticletitle{A closer look at memorization in deep networks}. In
  \bibinfo{booktitle}{\emph{International Conference on Machine Learning}}.
  \bibinfo{pages}{233--242}.
\newblock


\bibitem[\protect\citeauthoryear{Branson, Van~Horn, Belongie, and
  Perona}{Branson et~al\mbox{.}}{2014}]%
        {branson2014}
\bibfield{author}{\bibinfo{person}{Steve Branson}, \bibinfo{person}{Grant
  Van~Horn}, \bibinfo{person}{Serge Belongie}, {and} \bibinfo{person}{Pietro
  Perona}.} \bibinfo{year}{2014}\natexlab{}.
\newblock \showarticletitle{Bird species categorization using pose normalized
  deep convolutional nets}. In \bibinfo{booktitle}{\emph{British Machine Vision
  Conference}}.
\newblock


\bibitem[\protect\citeauthoryear{Chang, Learned-Miller, and McCallum}{Chang
  et~al\mbox{.}}{2017a}]%
        {activebias}
\bibfield{author}{\bibinfo{person}{Haw-Shiuan Chang}, \bibinfo{person}{Erik
  Learned-Miller}, {and} \bibinfo{person}{Andrew McCallum}.}
  \bibinfo{year}{2017}\natexlab{a}.
\newblock \showarticletitle{Active bias: Training more accurate neural networks
  by emphasizing high variance samples}. In \bibinfo{booktitle}{\emph{Advances
  in Neural Information Processing Systems}}. \bibinfo{pages}{1002--1012}.
\newblock


\bibitem[\protect\citeauthoryear{Chang, Learned-Miller, and McCallum}{Chang
  et~al\mbox{.}}{2017b}]%
        {chang2017active}
\bibfield{author}{\bibinfo{person}{Haw-Shiuan Chang}, \bibinfo{person}{Erik
  Learned-Miller}, {and} \bibinfo{person}{Andrew McCallum}.}
  \bibinfo{year}{2017}\natexlab{b}.
\newblock \showarticletitle{Active bias: Training more accurate neural networks
  by emphasizing high variance samples}. In \bibinfo{booktitle}{\emph{Advances
  in Neural Information Processing Systems}}. \bibinfo{pages}{1002--1012}.
\newblock


\bibitem[\protect\citeauthoryear{Chen, Zhang, Xie, Yao, Huang, and Tang}{Chen
  et~al\mbox{.}}{2020}]%
        {chen2020classification}
\bibfield{author}{\bibinfo{person}{Tao Chen}, \bibinfo{person}{Jian Zhang},
  \bibinfo{person}{Guo-Sen Xie}, \bibinfo{person}{Yazhou Yao},
  \bibinfo{person}{Xiaoshui Huang}, {and} \bibinfo{person}{Zhenmin Tang}.}
  \bibinfo{year}{2020}\natexlab{}.
\newblock \showarticletitle{Classification Constrained Discriminator For Domain
  Adaptive Semantic Segmentation}. In \bibinfo{booktitle}{\emph{IEEE
  International Conference on Multimedia and Expo}}. \bibinfo{pages}{1--6}.
\newblock


\bibitem[\protect\citeauthoryear{Chen, Bai, Zhang, and Mei}{Chen
  et~al\mbox{.}}{2019}]%
        {destructionconstruction2019}
\bibfield{author}{\bibinfo{person}{Yue Chen}, \bibinfo{person}{Yalong Bai},
  \bibinfo{person}{Wei Zhang}, {and} \bibinfo{person}{Tao Mei}.}
  \bibinfo{year}{2019}\natexlab{}.
\newblock \showarticletitle{Destruction and construction learning for
  fine-grained image recognition}. In \bibinfo{booktitle}{\emph{IEEE Conference
  on Computer Vision and Pattern Recognition}}. \bibinfo{pages}{5157--5166}.
\newblock


\bibitem[\protect\citeauthoryear{Cui, Song, Sun, Howard, and Belongie}{Cui
  et~al\mbox{.}}{2018}]%
        {cui2018large}
\bibfield{author}{\bibinfo{person}{Yin Cui}, \bibinfo{person}{Yang Song},
  \bibinfo{person}{Chen Sun}, \bibinfo{person}{Andrew Howard}, {and}
  \bibinfo{person}{Serge Belongie}.} \bibinfo{year}{2018}\natexlab{}.
\newblock \showarticletitle{Large scale fine-grained categorization and
  domain-specific transfer learning}. In \bibinfo{booktitle}{\emph{IEEE
  Conference on Computer Vision and Pattern Recognition}}.
  \bibinfo{pages}{4109--4118}.
\newblock


\bibitem[\protect\citeauthoryear{Cui, Zhou, Lin, and Belongie}{Cui
  et~al\mbox{.}}{2016}]%
        {cui2016fine}
\bibfield{author}{\bibinfo{person}{Yin Cui}, \bibinfo{person}{Feng Zhou},
  \bibinfo{person}{Yuanqing Lin}, {and} \bibinfo{person}{Serge Belongie}.}
  \bibinfo{year}{2016}\natexlab{}.
\newblock \showarticletitle{Fine-grained categorization and dataset
  bootstrapping using deep metric learning with humans in the loop}. In
  \bibinfo{booktitle}{\emph{IEEE Conference on Computer Vision and Pattern
  Recognition}}. \bibinfo{pages}{1153--1162}.
\newblock


\bibitem[\protect\citeauthoryear{Cui, Zhou, Wang, Liu, Lin, and Belongie}{Cui
  et~al\mbox{.}}{2017}]%
        {cui2017}
\bibfield{author}{\bibinfo{person}{Yin Cui}, \bibinfo{person}{Feng Zhou},
  \bibinfo{person}{Jiang Wang}, \bibinfo{person}{Xiao Liu},
  \bibinfo{person}{Yuanqing Lin}, {and} \bibinfo{person}{Serge Belongie}.}
  \bibinfo{year}{2017}\natexlab{}.
\newblock \showarticletitle{Kernel pooling for convolutional neural networks}.
  In \bibinfo{booktitle}{\emph{IEEE Conference on Computer Vision and Pattern
  Recognition}}. \bibinfo{pages}{2921--2930}.
\newblock


\bibitem[\protect\citeauthoryear{Dubey, Gupta, Raskar, and Naik}{Dubey
  et~al\mbox{.}}{2018}]%
        {dubey2018}
\bibfield{author}{\bibinfo{person}{Abhimanyu Dubey}, \bibinfo{person}{Otkrist
  Gupta}, \bibinfo{person}{Ramesh Raskar}, {and} \bibinfo{person}{Nikhil
  Naik}.} \bibinfo{year}{2018}\natexlab{}.
\newblock \showarticletitle{Maximum-entropy fine grained classification}. In
  \bibinfo{booktitle}{\emph{Advances in Neural Information Processing
  Systems}}. \bibinfo{pages}{637--647}.
\newblock


\bibitem[\protect\citeauthoryear{Gao, Beijbom, Zhang, and Darrell}{Gao
  et~al\mbox{.}}{2016}]%
        {gao2016}
\bibfield{author}{\bibinfo{person}{Yang Gao}, \bibinfo{person}{Oscar Beijbom},
  \bibinfo{person}{Ning Zhang}, {and} \bibinfo{person}{Trevor Darrell}.}
  \bibinfo{year}{2016}\natexlab{}.
\newblock \showarticletitle{Compact bilinear pooling}. In
  \bibinfo{booktitle}{\emph{IEEE Conference on Computer Vision and Pattern
  Recognition}}. \bibinfo{pages}{317--326}.
\newblock


\bibitem[\protect\citeauthoryear{Ge, Lin, and Yu}{Ge et~al\mbox{.}}{2019}]%
        {complementarypart2019}
\bibfield{author}{\bibinfo{person}{Weifeng Ge}, \bibinfo{person}{Xiangru Lin},
  {and} \bibinfo{person}{Yizhou Yu}.} \bibinfo{year}{2019}\natexlab{}.
\newblock \showarticletitle{Weakly supervised complementary parts models for
  fine-grained image classification from the bottom up}. In
  \bibinfo{booktitle}{\emph{IEEE Conference on Computer Vision and Pattern
  Recognition}}. \bibinfo{pages}{3034--3043}.
\newblock


\bibitem[\protect\citeauthoryear{Goldberger and Ben-Reuven}{Goldberger and
  Ben-Reuven}{2017}]%
        {goldberger2017}
\bibfield{author}{\bibinfo{person}{Jacob Goldberger} {and}
  \bibinfo{person}{Ehud Ben-Reuven}.} \bibinfo{year}{2017}\natexlab{}.
\newblock \showarticletitle{Training deep neural-networks using a noise
  adaptation layer}. In \bibinfo{booktitle}{\emph{International Conference on
  Learning Representations}}.
\newblock


\bibitem[\protect\citeauthoryear{Han, Yao, Yu, Niu, Xu, Hu, Tsang, and
  Sugiyama}{Han et~al\mbox{.}}{2018}]%
        {coteaching}
\bibfield{author}{\bibinfo{person}{Bo Han}, \bibinfo{person}{Quanming Yao},
  \bibinfo{person}{Xingrui Yu}, \bibinfo{person}{Gang Niu},
  \bibinfo{person}{Miao Xu}, \bibinfo{person}{Weihua Hu}, \bibinfo{person}{Ivor
  Tsang}, {and} \bibinfo{person}{Masashi Sugiyama}.}
  \bibinfo{year}{2018}\natexlab{}.
\newblock \showarticletitle{Co-teaching: Robust training of deep neural
  networks with extremely noisy labels}. In \bibinfo{booktitle}{\emph{Advances
  in Neural Information Processing Systems}}. \bibinfo{pages}{8527--8537}.
\newblock


\bibitem[\protect\citeauthoryear{He, Zhang, Ren, and Sun}{He
  et~al\mbox{.}}{2016}]%
        {resnet}
\bibfield{author}{\bibinfo{person}{Kaiming He}, \bibinfo{person}{Xiangyu
  Zhang}, \bibinfo{person}{Shaoqing Ren}, {and} \bibinfo{person}{Jian Sun}.}
  \bibinfo{year}{2016}\natexlab{}.
\newblock \showarticletitle{Deep residual learning for image recognition}. In
  \bibinfo{booktitle}{\emph{IEEE Conference on Computer Vision and Pattern
  Recognition}}. \bibinfo{pages}{770--778}.
\newblock


\bibitem[\protect\citeauthoryear{Hua, Shen, Zhang, and Tang}{Hua
  et~al\mbox{.}}{2016}]%
        {2016domain}
\bibfield{author}{\bibinfo{person}{Xian-sheng Hua}, \bibinfo{person}{Fumin
  Shen}, \bibinfo{person}{Jian Zhang}, {and} \bibinfo{person}{Zhenmin Tang}.}
  \bibinfo{year}{2016}\natexlab{}.
\newblock \showarticletitle{A domain robust approach for image dataset
  construction}. In \bibinfo{booktitle}{\emph{ACM international conference on
  Multimedia}}. \bibinfo{pages}{212--216}.
\newblock


\bibitem[\protect\citeauthoryear{Huang, Xu, Tao, and Zhang}{Huang
  et~al\mbox{.}}{2016}]%
        {huang2016}
\bibfield{author}{\bibinfo{person}{Shaoli Huang}, \bibinfo{person}{Zhe Xu},
  \bibinfo{person}{Dacheng Tao}, {and} \bibinfo{person}{Ya Zhang}.}
  \bibinfo{year}{2016}\natexlab{}.
\newblock \showarticletitle{Part-stacked cnn for fine-grained visual
  categorization}. In \bibinfo{booktitle}{\emph{IEEE Conference on Computer
  Vision and Pattern Recognition}}. \bibinfo{pages}{1173--1182}.
\newblock


\bibitem[\protect\citeauthoryear{Jiang, Zhou, Leung, Li, and Fei-Fei}{Jiang
  et~al\mbox{.}}{2017}]%
        {mentornet}
\bibfield{author}{\bibinfo{person}{Lu Jiang}, \bibinfo{person}{Zhengyuan Zhou},
  \bibinfo{person}{Thomas Leung}, \bibinfo{person}{Li-Jia Li}, {and}
  \bibinfo{person}{Li Fei-Fei}.} \bibinfo{year}{2017}\natexlab{}.
\newblock \showarticletitle{Mentornet: Learning data-driven curriculum for very
  deep neural networks on corrupted labels}. In
  \bibinfo{booktitle}{\emph{International Conference on Machine Learning}}.
\newblock


\bibitem[\protect\citeauthoryear{Kong and Fowlkes}{Kong and Fowlkes}{2017}]%
        {kong2017}
\bibfield{author}{\bibinfo{person}{Shu Kong} {and} \bibinfo{person}{Charless
  Fowlkes}.} \bibinfo{year}{2017}\natexlab{}.
\newblock \showarticletitle{Low-rank bilinear pooling for fine-grained
  classification}. In \bibinfo{booktitle}{\emph{IEEE Conference on Computer
  Vision and Pattern Recognition}}. \bibinfo{pages}{365--374}.
\newblock


\bibitem[\protect\citeauthoryear{Krause, Stark, Deng, and Fei-Fei}{Krause
  et~al\mbox{.}}{2013}]%
        {stanford-cars}
\bibfield{author}{\bibinfo{person}{Jonathan Krause}, \bibinfo{person}{Michael
  Stark}, \bibinfo{person}{Jia Deng}, {and} \bibinfo{person}{Li Fei-Fei}.}
  \bibinfo{year}{2013}\natexlab{}.
\newblock \showarticletitle{3d object representations for fine-grained
  categorization}. In \bibinfo{booktitle}{\emph{CVPRW}}.
  \bibinfo{pages}{554--561}.
\newblock


\bibitem[\protect\citeauthoryear{Krizhevsky and Hinton}{Krizhevsky and
  Hinton}{2009}]%
        {krizhevsky2009}
\bibfield{author}{\bibinfo{person}{Alex Krizhevsky} {and}
  \bibinfo{person}{Geoffrey Hinton}.} \bibinfo{year}{2009}\natexlab{}.
\newblock \showarticletitle{Learning multiple layers of features from tiny
  images}.
\newblock \bibinfo{journal}{\emph{Technical report, University of Toronto}}
  \bibinfo{volume}{1}, \bibinfo{number}{4} (\bibinfo{year}{2009}),
  \bibinfo{pages}{7}.
\newblock


\bibitem[\protect\citeauthoryear{Kumar, Packer, and Koller}{Kumar
  et~al\mbox{.}}{2010}]%
        {kumar2010}
\bibfield{author}{\bibinfo{person}{M~Pawan Kumar}, \bibinfo{person}{Benjamin
  Packer}, {and} \bibinfo{person}{Daphne Koller}.}
  \bibinfo{year}{2010}\natexlab{}.
\newblock \showarticletitle{Self-paced learning for latent variable models}. In
  \bibinfo{booktitle}{\emph{Advances in Neural Information Processing
  Systems}}. \bibinfo{pages}{1189--1197}.
\newblock


\bibitem[\protect\citeauthoryear{Lam, Mahasseni, and Todorovic}{Lam
  et~al\mbox{.}}{2017}]%
        {hsnet2017}
\bibfield{author}{\bibinfo{person}{Michael Lam}, \bibinfo{person}{Behrooz
  Mahasseni}, {and} \bibinfo{person}{Sinisa Todorovic}.}
  \bibinfo{year}{2017}\natexlab{}.
\newblock \showarticletitle{Fine-grained recognition as hsnet search for
  informative image parts}. In \bibinfo{booktitle}{\emph{IEEE Conference on
  Computer Vision and Pattern Recognition}}. \bibinfo{pages}{2520--2529}.
\newblock


\bibitem[\protect\citeauthoryear{Li, Xie, Wang, and Gao}{Li
  et~al\mbox{.}}{2018}]%
        {li2018}
\bibfield{author}{\bibinfo{person}{Peihua Li}, \bibinfo{person}{Jiangtao Xie},
  \bibinfo{person}{Qilong Wang}, {and} \bibinfo{person}{Zilin Gao}.}
  \bibinfo{year}{2018}\natexlab{}.
\newblock \showarticletitle{Towards faster training of global covariance
  pooling networks by iterative matrix square root normalization}. In
  \bibinfo{booktitle}{\emph{IEEE Conference on Computer Vision and Pattern
  Recognition}}. \bibinfo{pages}{947--955}.
\newblock


\bibitem[\protect\citeauthoryear{Lin, Goyal, Girshick, He, and Doll{\'a}r}{Lin
  et~al\mbox{.}}{2017a}]%
        {focalloss2017}
\bibfield{author}{\bibinfo{person}{Tsung-Yi Lin}, \bibinfo{person}{Priya
  Goyal}, \bibinfo{person}{Ross Girshick}, \bibinfo{person}{Kaiming He}, {and}
  \bibinfo{person}{Piotr Doll{\'a}r}.} \bibinfo{year}{2017}\natexlab{a}.
\newblock \showarticletitle{Focal loss for dense object detection}. In
  \bibinfo{booktitle}{\emph{IEEE International Conference on Computer Vision}}.
  \bibinfo{pages}{2980--2988}.
\newblock


\bibitem[\protect\citeauthoryear{Lin and Maji}{Lin and Maji}{2017}]%
        {linmaji2017}
\bibfield{author}{\bibinfo{person}{Tsung-Yu Lin} {and}
  \bibinfo{person}{Subhransu Maji}.} \bibinfo{year}{2017}\natexlab{}.
\newblock \showarticletitle{Improved bilinear pooling with cnns}. In
  \bibinfo{booktitle}{\emph{British Machine Vision Conference}}.
  \bibinfo{pages}{117.1--117.12}.
\newblock


\bibitem[\protect\citeauthoryear{Lin, RoyChowdhury, and Maji}{Lin
  et~al\mbox{.}}{2017b}]%
        {lin2017}
\bibfield{author}{\bibinfo{person}{Tsung-Yu Lin}, \bibinfo{person}{Aruni
  RoyChowdhury}, {and} \bibinfo{person}{Subhransu Maji}.}
  \bibinfo{year}{2017}\natexlab{b}.
\newblock \showarticletitle{Bilinear convolutional neural networks for
  fine-grained visual recognition}.
\newblock  \bibinfo{volume}{40}, \bibinfo{number}{6} (\bibinfo{year}{2017}),
  \bibinfo{pages}{1309--1322}.
\newblock


\bibitem[\protect\citeauthoryear{Lu, Liu, Yao, Tao, Tang, and Lu}{Lu
  et~al\mbox{.}}{2020}]%
        {lu2020hsi}
\bibfield{author}{\bibinfo{person}{Jiarou Lu}, \bibinfo{person}{Huafeng Liu},
  \bibinfo{person}{Yazhou Yao}, \bibinfo{person}{Shuyin Tao},
  \bibinfo{person}{Zhenming Tang}, {and} \bibinfo{person}{Jianfeng Lu}.}
  \bibinfo{year}{2020}\natexlab{}.
\newblock \showarticletitle{Hsi Road: A Hyper Spectral Image Dataset For Road
  Segmentation}. In \bibinfo{booktitle}{\emph{IEEE International Conference on
  Multimedia and Expo}}. \bibinfo{pages}{1--6}.
\newblock


\bibitem[\protect\citeauthoryear{Luo, Lin, Liu, Liu, Tang, and Yao}{Luo
  et~al\mbox{.}}{2019}]%
        {luo2019segeqa}
\bibfield{author}{\bibinfo{person}{Haonan Luo}, \bibinfo{person}{Guosheng Lin},
  \bibinfo{person}{Zichuan Liu}, \bibinfo{person}{Fayao Liu},
  \bibinfo{person}{Zhenmin Tang}, {and} \bibinfo{person}{Yazhou Yao}.}
  \bibinfo{year}{2019}\natexlab{}.
\newblock \showarticletitle{Segeqa: Video segmentation based visual attention
  for embodied question answering}. In \bibinfo{booktitle}{\emph{IEEE
  International Conference on Computer Vision}}. \bibinfo{pages}{9667--9676}.
\newblock


\bibitem[\protect\citeauthoryear{Maji, Rahtu, Kannala, Blaschko, and
  Vedaldi}{Maji et~al\mbox{.}}{2013}]%
        {fgvc-aircraft}
\bibfield{author}{\bibinfo{person}{Subhransu Maji}, \bibinfo{person}{Esa
  Rahtu}, \bibinfo{person}{Juho Kannala}, \bibinfo{person}{Matthew Blaschko},
  {and} \bibinfo{person}{Andrea Vedaldi}.} \bibinfo{year}{2013}\natexlab{}.
\newblock \showarticletitle{Fine-grained visual classification of aircraft}.
\newblock \bibinfo{journal}{\emph{arXiv:1306.5151}} (\bibinfo{year}{2013}).
\newblock


\bibitem[\protect\citeauthoryear{Malach and Shalev-Shwartz}{Malach and
  Shalev-Shwartz}{2017}]%
        {decoupling}
\bibfield{author}{\bibinfo{person}{Eran Malach} {and} \bibinfo{person}{Shai
  Shalev-Shwartz}.} \bibinfo{year}{2017}\natexlab{}.
\newblock \showarticletitle{Decoupling "when to update" from "how to update"}.
  In \bibinfo{booktitle}{\emph{Advances in Neural Information Processing
  Systems}}. \bibinfo{pages}{960--970}.
\newblock


\bibitem[\protect\citeauthoryear{Niu, Veeraraghavan, and Sabharwal}{Niu
  et~al\mbox{.}}{2018}]%
        {niu2018webly}
\bibfield{author}{\bibinfo{person}{Li Niu}, \bibinfo{person}{Ashok
  Veeraraghavan}, {and} \bibinfo{person}{Ashutosh Sabharwal}.}
  \bibinfo{year}{2018}\natexlab{}.
\newblock \showarticletitle{Webly supervised learning meets zero-shot learning:
  A hybrid approach for fine-grained classification}. In
  \bibinfo{booktitle}{\emph{IEEE Conference on Computer Vision and Pattern
  Recognition}}. \bibinfo{pages}{7171--7180}.
\newblock


\bibitem[\protect\citeauthoryear{Patrini, Rozza, Krishna~Menon, Nock, and
  Qu}{Patrini et~al\mbox{.}}{2017}]%
        {fcorrection}
\bibfield{author}{\bibinfo{person}{Giorgio Patrini},
  \bibinfo{person}{Alessandro Rozza}, \bibinfo{person}{Aditya Krishna~Menon},
  \bibinfo{person}{Richard Nock}, {and} \bibinfo{person}{Lizhen Qu}.}
  \bibinfo{year}{2017}\natexlab{}.
\newblock \showarticletitle{Making deep neural networks robust to label noise:
  A loss correction approach}. In \bibinfo{booktitle}{\emph{IEEE Conference on
  Computer Vision and Pattern Recognition}}. \bibinfo{pages}{1944--1952}.
\newblock


\bibitem[\protect\citeauthoryear{Reed, Lee, Anguelov, Szegedy, Erhan, and
  Rabinovich}{Reed et~al\mbox{.}}{2014}]%
        {bootstrap}
\bibfield{author}{\bibinfo{person}{Scott Reed}, \bibinfo{person}{Honglak Lee},
  \bibinfo{person}{Dragomir Anguelov}, \bibinfo{person}{Christian Szegedy},
  \bibinfo{person}{Dumitru Erhan}, {and} \bibinfo{person}{Andrew Rabinovich}.}
  \bibinfo{year}{2014}\natexlab{}.
\newblock \showarticletitle{Training deep neural networks on noisy labels with
  bootstrapping}. In \bibinfo{booktitle}{\emph{International Conference on
  Learning Representations}}. \bibinfo{pages}{1--11}.
\newblock


\bibitem[\protect\citeauthoryear{Ren, Zeng, Yang, and Urtasun}{Ren
  et~al\mbox{.}}{2018}]%
        {ren2018}
\bibfield{author}{\bibinfo{person}{Mengye Ren}, \bibinfo{person}{Wenyuan Zeng},
  \bibinfo{person}{Bin Yang}, {and} \bibinfo{person}{Raquel Urtasun}.}
  \bibinfo{year}{2018}\natexlab{}.
\newblock \showarticletitle{Learning to reweight examples for robust deep
  learning}. In \bibinfo{booktitle}{\emph{International Conference on Machine
  Learning}}. \bibinfo{pages}{4334--4343}.
\newblock


\bibitem[\protect\citeauthoryear{Shrivastava, Gupta, and Girshick}{Shrivastava
  et~al\mbox{.}}{2016}]%
        {harddatamining2016}
\bibfield{author}{\bibinfo{person}{Abhinav Shrivastava},
  \bibinfo{person}{Abhinav Gupta}, {and} \bibinfo{person}{Ross Girshick}.}
  \bibinfo{year}{2016}\natexlab{}.
\newblock \showarticletitle{Training region-based object detectors with online
  hard example mining}. In \bibinfo{booktitle}{\emph{IEEE Conference on
  Computer Vision and Pattern Recognition}}. \bibinfo{pages}{761--769}.
\newblock


\bibitem[\protect\citeauthoryear{Shu, Tang, Qi, Liu, and Yang}{Shu
  et~al\mbox{.}}{2019}]%
        {shu2019hierarchical}
\bibfield{author}{\bibinfo{person}{Xiangbo Shu}, \bibinfo{person}{Jinhui Tang},
  \bibinfo{person}{Guojun Qi}, \bibinfo{person}{Wei Liu}, {and}
  \bibinfo{person}{Jian Yang}.} \bibinfo{year}{2019}\natexlab{}.
\newblock \showarticletitle{Hierarchical long short-term concurrent memory for
  human interaction recognition}.
\newblock \bibinfo{journal}{\emph{IEEE transactions on pattern analysis and
  machine intelligence}} (\bibinfo{year}{2019}).
\newblock


\bibitem[\protect\citeauthoryear{Simonyan and Zisserman}{Simonyan and
  Zisserman}{2014}]%
        {vgg}
\bibfield{author}{\bibinfo{person}{Karen Simonyan} {and}
  \bibinfo{person}{Andrew Zisserman}.} \bibinfo{year}{2014}\natexlab{}.
\newblock \showarticletitle{Very deep convolutional networks for large-scale
  image recognition}.
\newblock \bibinfo{journal}{\emph{arXiv:1409.1556}} (\bibinfo{year}{2014}).
\newblock


\bibitem[\protect\citeauthoryear{Song, Kim, and Lee}{Song
  et~al\mbox{.}}{2019}]%
        {selfie2019}
\bibfield{author}{\bibinfo{person}{Hwanjun Song}, \bibinfo{person}{Minseok
  Kim}, {and} \bibinfo{person}{Jae-Gil Lee}.} \bibinfo{year}{2019}\natexlab{}.
\newblock \showarticletitle{SELFIE: Refurbishing unclean samples for robust
  deep learning}. In \bibinfo{booktitle}{\emph{International Conference on
  Machine Learning}}. \bibinfo{pages}{5907--5915}.
\newblock


\bibitem[\protect\citeauthoryear{Sun, Shen, Liu, and Wang}{Sun
  et~al\mbox{.}}{2019}]%
        {yao2019dynamically}
\bibfield{author}{\bibinfo{person}{Zeren Sun}, \bibinfo{person}{Fumin Shen},
  \bibinfo{person}{Li Liu}, {and} \bibinfo{person}{Limin et~al. Wang}.}
  \bibinfo{year}{2019}\natexlab{}.
\newblock \showarticletitle{Dynamically visual disambiguation of keyword-based
  image search}.
\newblock \bibinfo{journal}{\emph{International Joint Conference on Artificial
  Intelligence}} (\bibinfo{year}{2019}), \bibinfo{pages}{996--1002}.
\newblock


\bibitem[\protect\citeauthoryear{Szegedy, Vanhoucke, Ioffe, Shlens, and
  Wojna}{Szegedy et~al\mbox{.}}{2016}]%
        {szegedy2016}
\bibfield{author}{\bibinfo{person}{Christian Szegedy}, \bibinfo{person}{Vincent
  Vanhoucke}, \bibinfo{person}{Sergey Ioffe}, \bibinfo{person}{Jon Shlens},
  {and} \bibinfo{person}{Zbigniew Wojna}.} \bibinfo{year}{2016}\natexlab{}.
\newblock \showarticletitle{Rethinking the inception architecture for computer
  vision}. In \bibinfo{booktitle}{\emph{IEEE Conference on Computer Vision and
  Pattern Recognition}}. \bibinfo{pages}{2818--2826}.
\newblock


\bibitem[\protect\citeauthoryear{Tang, Li, Lai, Zhang, Yan, et~al\mbox{.}}{Tang
  et~al\mbox{.}}{2017}]%
        {tang2017personalized}
\bibfield{author}{\bibinfo{person}{Jinhui Tang}, \bibinfo{person}{Zechao Li},
  \bibinfo{person}{Hanjiang Lai}, \bibinfo{person}{Liyan Zhang},
  \bibinfo{person}{Shuicheng Yan}, {et~al\mbox{.}}}
  \bibinfo{year}{2017}\natexlab{}.
\newblock \showarticletitle{Personalized age progression with bi-level aging
  dictionary learning}.
\newblock \bibinfo{journal}{\emph{IEEE transactions on pattern analysis and
  machine intelligence}} \bibinfo{volume}{40}, \bibinfo{number}{4}
  (\bibinfo{year}{2017}), \bibinfo{pages}{905--917}.
\newblock


\bibitem[\protect\citeauthoryear{Van~Horn, Mac~Aodha, Song, Cui, Sun, Shepard,
  Adam, Perona, and Belongie}{Van~Horn et~al\mbox{.}}{2018}]%
        {inat17}
\bibfield{author}{\bibinfo{person}{Grant Van~Horn}, \bibinfo{person}{Oisin
  Mac~Aodha}, \bibinfo{person}{Yang Song}, \bibinfo{person}{Yin Cui},
  \bibinfo{person}{Chen Sun}, \bibinfo{person}{Alex Shepard},
  \bibinfo{person}{Hartwig Adam}, \bibinfo{person}{Pietro Perona}, {and}
  \bibinfo{person}{Serge Belongie}.} \bibinfo{year}{2018}\natexlab{}.
\newblock \showarticletitle{The inaturalist species classification and
  detection dataset}. In \bibinfo{booktitle}{\emph{IEEE Conference on Computer
  Vision and Pattern Recognition}}. \bibinfo{pages}{8769--8778}.
\newblock


\bibitem[\protect\citeauthoryear{Wah, Branson, Welinder, Perona, and
  Belongie}{Wah et~al\mbox{.}}{2011}]%
        {cub200-2011}
\bibfield{author}{\bibinfo{person}{Catherine Wah}, \bibinfo{person}{Steve
  Branson}, \bibinfo{person}{Peter Welinder}, \bibinfo{person}{Pietro Perona},
  {and} \bibinfo{person}{Serge Belongie}.} \bibinfo{year}{2011}\natexlab{}.
\newblock \showarticletitle{The Caltech-UCSD Birds-200-2011 Dataset.}
\newblock \bibinfo{journal}{\emph{CNS-TR-2011-001}} (\bibinfo{year}{2011}).
\newblock


\bibitem[\protect\citeauthoryear{Wei, Xie, Wu, and Shen}{Wei
  et~al\mbox{.}}{2018}]%
        {wei2018}
\bibfield{author}{\bibinfo{person}{Xiu-Shen Wei}, \bibinfo{person}{Chen-Wei
  Xie}, \bibinfo{person}{Jianxin Wu}, {and} \bibinfo{person}{Chunhua Shen}.}
  \bibinfo{year}{2018}\natexlab{}.
\newblock \showarticletitle{Mask-CNN: Localizing parts and selecting
  descriptors for fine-grained bird species categorization}.
\newblock \bibinfo{journal}{\emph{Pattern Recognition}}  \bibinfo{volume}{76}
  (\bibinfo{year}{2018}), \bibinfo{pages}{704--714}.
\newblock


\bibitem[\protect\citeauthoryear{Xie, Liu, Jin, Zhu, Zhang, Qin, Yao, and
  Shao}{Xie et~al\mbox{.}}{2019}]%
        {xie2019attentive}
\bibfield{author}{\bibinfo{person}{Guo-Sen Xie}, \bibinfo{person}{Li Liu},
  \bibinfo{person}{Xiaobo Jin}, \bibinfo{person}{Fan Zhu},
  \bibinfo{person}{Zheng Zhang}, \bibinfo{person}{Jie Qin},
  \bibinfo{person}{Yazhou Yao}, {and} \bibinfo{person}{Ling Shao}.}
  \bibinfo{year}{2019}\natexlab{}.
\newblock \showarticletitle{Attentive region embedding network for zero-shot
  learning}. In \bibinfo{booktitle}{\emph{IEEE Conference on Computer Vision
  and Pattern Recognition}}. \bibinfo{pages}{9384--9393}.
\newblock


\bibitem[\protect\citeauthoryear{Xie, Liu, Zhu, Zhao, Zhang, Yao, Qin, and
  Shao}{Xie et~al\mbox{.}}{2020}]%
        {xie2020eccv}
\bibfield{author}{\bibinfo{person}{Guo-Sen Xie}, \bibinfo{person}{Li Liu},
  \bibinfo{person}{Fan Zhu}, \bibinfo{person}{Fang Zhao},
  \bibinfo{person}{Zheng Zhang}, \bibinfo{person}{Yazhou Yao},
  \bibinfo{person}{Jie Qin}, {and} \bibinfo{person}{Ling Shao}.}
  \bibinfo{year}{2020}\natexlab{}.
\newblock \showarticletitle{Region Graph Embedding Network for Zero-Shot
  Learning}. In \bibinfo{booktitle}{\emph{European Conference on Computer
  Vision}}.
\newblock


\bibitem[\protect\citeauthoryear{Xu, Huang, Zhang, and Tao}{Xu
  et~al\mbox{.}}{2016}]%
        {xu2018webly}
\bibfield{author}{\bibinfo{person}{Zhe Xu}, \bibinfo{person}{Shaoli Huang},
  \bibinfo{person}{Ya Zhang}, {and} \bibinfo{person}{Dacheng Tao}.}
  \bibinfo{year}{2016}\natexlab{}.
\newblock \showarticletitle{Webly-supervised fine-grained visual categorization
  via deep domain adaptation}.
\newblock  \bibinfo{volume}{40}, \bibinfo{number}{5} (\bibinfo{year}{2016}),
  \bibinfo{pages}{1100--1113}.
\newblock


\bibitem[\protect\citeauthoryear{Yang, Sun, Lai, Zheng, and Cheng}{Yang
  et~al\mbox{.}}{2018}]%
        {yang2018recognition}
\bibfield{author}{\bibinfo{person}{Jufeng Yang}, \bibinfo{person}{Xiaoxiao
  Sun}, \bibinfo{person}{Yu-Kun Lai}, \bibinfo{person}{Liang Zheng}, {and}
  \bibinfo{person}{Ming-Ming Cheng}.} \bibinfo{year}{2018}\natexlab{}.
\newblock \showarticletitle{Recognition from web data: A progressive filtering
  approach}.
\newblock  \bibinfo{volume}{27}, \bibinfo{number}{11} (\bibinfo{year}{2018}),
  \bibinfo{pages}{5303--5315}.
\newblock


\bibitem[\protect\citeauthoryear{Yao, Zhang, Zhang, Li, and Tian}{Yao
  et~al\mbox{.}}{2016}]%
        {coarse-to-fine}
\bibfield{author}{\bibinfo{person}{Hantao Yao}, \bibinfo{person}{Shiliang
  Zhang}, \bibinfo{person}{Yongdong Zhang}, \bibinfo{person}{Jintao Li}, {and}
  \bibinfo{person}{Qi Tian}.} \bibinfo{year}{2016}\natexlab{}.
\newblock \showarticletitle{Coarse-to-fine description for fine-grained visual
  categorization}.
\newblock  \bibinfo{volume}{25}, \bibinfo{number}{10} (\bibinfo{year}{2016}),
  \bibinfo{pages}{4858--4872}.
\newblock


\bibitem[\protect\citeauthoryear{Yao, Shen, Xie, Liu, Zhu, Zhang, and Shen}{Yao
  et~al\mbox{.}}{2020}]%
        {yao2020exploiting}
\bibfield{author}{\bibinfo{person}{Yazhou Yao}, \bibinfo{person}{Fumin Shen},
  \bibinfo{person}{Guosen Xie}, \bibinfo{person}{Li Liu}, \bibinfo{person}{Fan
  Zhu}, \bibinfo{person}{Jian Zhang}, {and} \bibinfo{person}{Heng~Tao Shen}.}
  \bibinfo{year}{2020}\natexlab{}.
\newblock \showarticletitle{Exploiting web images for multi-output
  classification: From category to subcategories}.
\newblock \bibinfo{journal}{\emph{IEEE Transactions on Neural Networks and
  Learning Systems}} \bibinfo{volume}{31}, \bibinfo{number}{7}
  (\bibinfo{year}{2020}), \bibinfo{pages}{2348--2360}.
\newblock


\bibitem[\protect\citeauthoryear{Yao, Shen, Zhang, Liu, Tang, and Shao}{Yao
  et~al\mbox{.}}{2018a}]%
        {yao2018extracting}
\bibfield{author}{\bibinfo{person}{Yazhou Yao}, \bibinfo{person}{Fumin Shen},
  \bibinfo{person}{Jian Zhang}, \bibinfo{person}{Li Liu},
  \bibinfo{person}{Zhenmin Tang}, {and} \bibinfo{person}{Ling Shao}.}
  \bibinfo{year}{2018}\natexlab{a}.
\newblock \showarticletitle{Extracting multiple visual senses for web
  learning}.
\newblock \bibinfo{journal}{\emph{IEEE Transactions on Multimedia}}
  \bibinfo{volume}{21}, \bibinfo{number}{1} (\bibinfo{year}{2018}),
  \bibinfo{pages}{184--196}.
\newblock


\bibitem[\protect\citeauthoryear{Yao, Shen, Zhang, Liu, Tang, and Shao}{Yao
  et~al\mbox{.}}{2018b}]%
        {yao2018tip}
\bibfield{author}{\bibinfo{person}{Yazhou Yao}, \bibinfo{person}{Fumin Shen},
  \bibinfo{person}{Jian Zhang}, \bibinfo{person}{Li Liu},
  \bibinfo{person}{Zhenmin Tang}, {and} \bibinfo{person}{Ling Shao}.}
  \bibinfo{year}{2018}\natexlab{b}.
\newblock \showarticletitle{Extracting privileged information for enhancing
  classifier learning}.
\newblock \bibinfo{journal}{\emph{IEEE Transactions on Image Processing}}
  \bibinfo{volume}{28}, \bibinfo{number}{1} (\bibinfo{year}{2018}),
  \bibinfo{pages}{436--450}.
\newblock


\bibitem[\protect\citeauthoryear{Yao, Zhang, Shen, Hua, Xu, and Tang}{Yao
  et~al\mbox{.}}{2017}]%
        {yao2017exploiting}
\bibfield{author}{\bibinfo{person}{Yazhou Yao}, \bibinfo{person}{Jian Zhang},
  \bibinfo{person}{Fumin Shen}, \bibinfo{person}{Xiansheng Hua},
  \bibinfo{person}{Jingsong Xu}, {and} \bibinfo{person}{Zhenmin Tang}.}
  \bibinfo{year}{2017}\natexlab{}.
\newblock \showarticletitle{Exploiting web images for dataset construction: A
  domain robust approach}.
\newblock \bibinfo{journal}{\emph{IEEE Transactions on Multimedia}}
  \bibinfo{volume}{19}, \bibinfo{number}{8} (\bibinfo{year}{2017}),
  \bibinfo{pages}{1771--1784}.
\newblock


\bibitem[\protect\citeauthoryear{Yao, Zhang, Shen, Liu, Zhu, Zhang, and
  Shen}{Yao et~al\mbox{.}}{2019}]%
        {yao2019towards}
\bibfield{author}{\bibinfo{person}{Yazhou Yao}, \bibinfo{person}{Jian Zhang},
  \bibinfo{person}{Fumin Shen}, \bibinfo{person}{Li Liu}, \bibinfo{person}{Fan
  Zhu}, \bibinfo{person}{Dongxiang Zhang}, {and} \bibinfo{person}{Heng~Tao
  Shen}.} \bibinfo{year}{2019}\natexlab{}.
\newblock \showarticletitle{Towards automatic construction of diverse,
  high-quality image datasets}.
\newblock \bibinfo{journal}{\emph{IEEE Transactions on Knowledge and Data
  Engineering}} \bibinfo{volume}{32}, \bibinfo{number}{6}
  (\bibinfo{year}{2019}), \bibinfo{pages}{1199--1211}.
\newblock


\bibitem[\protect\citeauthoryear{Yi and Wu}{Yi and Wu}{2019}]%
        {pencil2019}
\bibfield{author}{\bibinfo{person}{Kun Yi} {and} \bibinfo{person}{Jianxin Wu}.}
  \bibinfo{year}{2019}\natexlab{}.
\newblock \showarticletitle{Probabilistic end-to-end noise correction for
  learning with noisy labels}. In \bibinfo{booktitle}{\emph{IEEE Conference on
  Computer Vision and Pattern Recognition}}. \bibinfo{pages}{7017--7025}.
\newblock


\bibitem[\protect\citeauthoryear{Zhang, Bengio, Hardt, Recht, and
  Vinyals}{Zhang et~al\mbox{.}}{2017a}]%
        {zhang2016understanding}
\bibfield{author}{\bibinfo{person}{Chiyuan Zhang}, \bibinfo{person}{Samy
  Bengio}, \bibinfo{person}{Moritz Hardt}, \bibinfo{person}{Benjamin Recht},
  {and} \bibinfo{person}{Oriol Vinyals}.} \bibinfo{year}{2017}\natexlab{a}.
\newblock \showarticletitle{Understanding deep learning requires rethinking
  generalization}. In \bibinfo{booktitle}{\emph{International Conference on
  Learning Representations}}.
\newblock


\bibitem[\protect\citeauthoryear{Zhang, Yao, Liu, Xie, Shu, Zhou, Zhang, Shen,
  and Tang}{Zhang et~al\mbox{.}}{2020a}]%
        {aaai20}
\bibfield{author}{\bibinfo{person}{Chuanyi Zhang}, \bibinfo{person}{Yazhou
  Yao}, \bibinfo{person}{Huafeng Liu}, \bibinfo{person}{Guo-Sen Xie},
  \bibinfo{person}{Xiangbo Shu}, \bibinfo{person}{Tianfei Zhou},
  \bibinfo{person}{Zheng Zhang}, \bibinfo{person}{Fumin Shen}, {and}
  \bibinfo{person}{Zhenmin Tang}.} \bibinfo{year}{2020}\natexlab{a}.
\newblock \showarticletitle{Web-Supervised Network with Softly Update-Drop
  Training for Fine-Grained Visual Classification}. In
  \bibinfo{booktitle}{\emph{AAAI Conference on Artificial Intelligence}}.
  \bibinfo{pages}{12781--12788}.
\newblock


\bibitem[\protect\citeauthoryear{Zhang, Yao, Zhang, Chen, and et~al.}{Zhang
  et~al\mbox{.}}{2020b}]%
        {zhang2020web}
\bibfield{author}{\bibinfo{person}{Chuanyi Zhang}, \bibinfo{person}{Yazhou
  Yao}, \bibinfo{person}{Jiachao Zhang}, \bibinfo{person}{Jiaxin Chen}, {and}
  \bibinfo{person}{et al.}} \bibinfo{year}{2020}\natexlab{b}.
\newblock \showarticletitle{Web-Supervised Network for Fine-Grained Visual
  Classification}. In \bibinfo{booktitle}{\emph{IEEE International Conference
  on Multimedia and Expo}}. \bibinfo{pages}{1--6}.
\newblock


\bibitem[\protect\citeauthoryear{Zhang, Shen, Hua, Xu, and Tang}{Zhang
  et~al\mbox{.}}{2016}]%
        {2016automatic}
\bibfield{author}{\bibinfo{person}{Jian Zhang}, \bibinfo{person}{Fumin Shen},
  \bibinfo{person}{Xiansheng Hua}, \bibinfo{person}{Jingsong Xu}, {and}
  \bibinfo{person}{Zhenmin Tang}.} \bibinfo{year}{2016}\natexlab{}.
\newblock \showarticletitle{Automatic image dataset construction with multiple
  textual metadata}. In \bibinfo{booktitle}{\emph{IEEE International Conference
  on Multimedia and Expo}}. \bibinfo{pages}{1--6}.
\newblock


\bibitem[\protect\citeauthoryear{Zhang, Shen, Hua, Xu, and Tang}{Zhang
  et~al\mbox{.}}{2017b}]%
        {2017new}
\bibfield{author}{\bibinfo{person}{Jian Zhang}, \bibinfo{person}{Fumin Shen},
  \bibinfo{person}{Xiansheng Hua}, \bibinfo{person}{Jingsong Xu}, {and}
  \bibinfo{person}{Zhenmin Tang}.} \bibinfo{year}{2017}\natexlab{b}.
\newblock \showarticletitle{A new web-supervised method for image dataset
  constructions}.
\newblock \bibinfo{journal}{\emph{Neurocomputing}}  \bibinfo{volume}{236}
  (\bibinfo{year}{2017}), \bibinfo{pages}{23--31}.
\newblock


\bibitem[\protect\citeauthoryear{Zhang, Shen, Yang, Huang, and Tang}{Zhang
  et~al\mbox{.}}{2018}]%
        {2018discovering}
\bibfield{author}{\bibinfo{person}{Jian Zhang}, \bibinfo{person}{Fumin Shen},
  \bibinfo{person}{Wankou Yang}, \bibinfo{person}{Pu Huang}, {and}
  \bibinfo{person}{Zhenmin Tang}.} \bibinfo{year}{2018}\natexlab{}.
\newblock \showarticletitle{Discovering and distinguishing multiple visual
  senses for polysemous words}. In \bibinfo{booktitle}{\emph{AAAI Conference on
  Artificial Intelligence}}. \bibinfo{pages}{523--530}.
\newblock


\bibitem[\protect\citeauthoryear{Zheng, Fu, Zha, and Luo}{Zheng
  et~al\mbox{.}}{2019}]%
        {trilinear2019}
\bibfield{author}{\bibinfo{person}{Heliang Zheng}, \bibinfo{person}{Jianlong
  Fu}, \bibinfo{person}{Zheng-Jun Zha}, {and} \bibinfo{person}{Jiebo Luo}.}
  \bibinfo{year}{2019}\natexlab{}.
\newblock \showarticletitle{Looking for the devil in the details: Learning
  trilinear attention sampling network for fine-grained image recognition}. In
  \bibinfo{booktitle}{\emph{IEEE Conference on Computer Vision and Pattern
  Recognition}}. \bibinfo{pages}{5012--5021}.
\newblock


\bibitem[\protect\citeauthoryear{Zhou, Khosla, Lapedriza, Torralba, and
  Oliva}{Zhou et~al\mbox{.}}{2016}]%
        {pcaduplicateremoval2016}
\bibfield{author}{\bibinfo{person}{Bolei Zhou}, \bibinfo{person}{Aditya
  Khosla}, \bibinfo{person}{Agata Lapedriza}, \bibinfo{person}{Antonio
  Torralba}, {and} \bibinfo{person}{Aude Oliva}.}
  \bibinfo{year}{2016}\natexlab{}.
\newblock \showarticletitle{Places: An image database for deep scene
  understanding}.
\newblock \bibinfo{journal}{\emph{arXiv:1610.02055}} (\bibinfo{year}{2016}).
\newblock


\end{thebibliography}


\begin{thebibliography}{64}
	
	\ifx \showCODEN    \undefined \def \showCODEN     #1{\unskip}     \fi
	\ifx \showDOI      \undefined \def \showDOI       #1{#1}\fi
	\ifx \showISBNx    \undefined \def \showISBNx     #1{\unskip}     \fi
	\ifx \showISBNxiii \undefined \def \showISBNxiii  #1{\unskip}     \fi
	\ifx \showISSN     \undefined \def \showISSN      #1{\unskip}     \fi
	\ifx \showLCCN     \undefined \def \showLCCN      #1{\unskip}     \fi
	\ifx \shownote     \undefined \def \shownote      #1{#1}          \fi
	\ifx \showarticletitle \undefined \def \showarticletitle #1{#1}   \fi
	\ifx \showURL      \undefined \def \showURL       {\relax}        \fi
	\providecommand\bibfield[2]{#2}
	\providecommand\bibinfo[2]{#2}
	\providecommand\natexlab[1]{#1}
	\providecommand\showeprint[2][]{arXiv:#2}
	
	\bibitem[\protect\citeauthoryear{Arpit, Jastrz{\k{e}}bski, Ballas, Krueger,
		Bengio, Kanwal, Maharaj, Fischer, Courville, Bengio, et~al\mbox{.}}{Arpit
		et~al\mbox{.}}{2017}]%
	{motivation2017}
	\bibfield{author}{\bibinfo{person}{Devansh Arpit} et al.}
	\bibinfo{year}{2017}\natexlab{}.
	\newblock \showarticletitle{A closer look at memorization in deep networks}. In
	\bibinfo{booktitle}{\emph{ICML}}.
	\bibinfo{pages}{233--242}.
	\newblock
	
	
	\bibitem[\protect\citeauthoryear{Branson, Van~Horn, Belongie, and
		Perona}{Branson et~al\mbox{.}}{2014}]%
	{branson2014}
	\bibfield{author}{\bibinfo{person}{Steve Branson} et al.} \bibinfo{year}{2014}\natexlab{}.
	\newblock \showarticletitle{Bird species categorization using pose normalized
		deep convolutional nets}. In \bibinfo{booktitle}{\emph{BMVC}}.
	\newblock
	
	
	\bibitem[\protect\citeauthoryear{Chang, Learned-Miller, and McCallum}{Chang
		et~al\mbox{.}}{2017a}]%
	{activebias}
	\bibfield{author}{\bibinfo{person}{Haw-Shiuan Chang} et al.}
	\bibinfo{year}{2017}\natexlab{a}.
	\newblock \showarticletitle{Active bias: Training more accurate neural networks
		by emphasizing high variance samples}. In \bibinfo{booktitle}{\emph{NIPS}}. \bibinfo{pages}{1002--1012}.
	\newblock
	
	
	\bibitem[\protect\citeauthoryear{Chang, Learned-Miller, and McCallum}{Chang
		et~al\mbox{.}}{2017b}]%
	{chang2017active}
	\bibfield{author}{\bibinfo{person}{Haw-Shiuan Chang} et al.}
	\bibinfo{year}{2017}\natexlab{b}.
	\newblock \showarticletitle{Active bias: Training more accurate neural networks
		by emphasizing high variance samples}. In \bibinfo{booktitle}{\emph{NIPS}}. \bibinfo{pages}{1002--1012}.
	\newblock
	
	
	\bibitem[\protect\citeauthoryear{Chen, Zhang, Xie, Yao, Huang, and Tang}{Chen
		et~al\mbox{.}}{2020}]%
	{chen2020classification}
	\bibfield{author}{\bibinfo{person}{Tao Chen} et al.}
	\bibinfo{year}{2020}\natexlab{}.
	\newblock \showarticletitle{Classification Constrained Discriminator For Domain
		Adaptive Semantic Segmentation}. In \bibinfo{booktitle}{\emph{ICME}}. \bibinfo{pages}{1--6}.
	\newblock
	
	
	\bibitem[\protect\citeauthoryear{Chen, Bai, Zhang, and Mei}{Chen
		et~al\mbox{.}}{2019}]%
	{destructionconstruction2019}
	\bibfield{author}{\bibinfo{person}{Yue Chen} et al.}
	\bibinfo{year}{2019}\natexlab{}.
	\newblock \showarticletitle{Destruction and construction learning for
		fine-grained image recognition}. In \bibinfo{booktitle}{\emph{CVPR}}. \bibinfo{pages}{5157--5166}.
	\newblock
	
	
	\bibitem[\protect\citeauthoryear{Cui, Song, Sun, Howard, and Belongie}{Cui
		et~al\mbox{.}}{2018}]%
	{cui2018large}
	\bibfield{author}{\bibinfo{person}{Yin Cui} et al.} \bibinfo{year}{2018}\natexlab{}.
	\newblock \showarticletitle{Large scale fine-grained categorization and
		domain-specific transfer learning}. In \bibinfo{booktitle}{\emph{CVPR}}.
	\bibinfo{pages}{4109--4118}.
	\newblock
	
	
	\bibitem[\protect\citeauthoryear{Cui, Zhou, Lin, and Belongie}{Cui
		et~al\mbox{.}}{2016}]%
	{cui2016fine}
	\bibfield{author}{\bibinfo{person}{Yin Cui} et al.}
	\bibinfo{year}{2016}\natexlab{}.
	\newblock \showarticletitle{Fine-grained categorization and dataset
		bootstrapping using deep metric learning with humans in the loop}. In
	\bibinfo{booktitle}{\emph{CVPR}}. \bibinfo{pages}{1153--1162}.
	\newblock
	
	
	\bibitem[\protect\citeauthoryear{Cui, Zhou, Wang, Liu, Lin, and Belongie}{Cui
		et~al\mbox{.}}{2017}]%
	{cui2017}
	\bibfield{author}{\bibinfo{person}{Yin Cui} et al.}
	\bibinfo{year}{2017}\natexlab{}.
	\newblock \showarticletitle{Kernel pooling for convolutional neural networks}.
	In \bibinfo{booktitle}{\emph{CVPR}}. \bibinfo{pages}{2921--2930}.
	\newblock
	
	
	\bibitem[\protect\citeauthoryear{Dubey, Gupta, Raskar, and Naik}{Dubey
		et~al\mbox{.}}{2018}]%
	{dubey2018}
	\bibfield{author}{\bibinfo{person}{Abhimanyu Dubey} et al.} \bibinfo{year}{2018}\natexlab{}.
	\newblock \showarticletitle{Maximum-entropy fine grained classification}. In
	\bibinfo{booktitle}{\emph{NIPS}}. \bibinfo{pages}{637--647}.
	\newblock
	
	
	\bibitem[\protect\citeauthoryear{Gao, Beijbom, Zhang, and Darrell}{Gao
		et~al\mbox{.}}{2016}]%
	{gao2016}
	\bibfield{author}{\bibinfo{person}{Yang Gao} et al.}
	\bibinfo{year}{2016}\natexlab{}.
	\newblock \showarticletitle{Compact bilinear pooling}. In
	\bibinfo{booktitle}{\emph{CVPR}}. \bibinfo{pages}{317--326}.
	\newblock
	
	
	\bibitem[\protect\citeauthoryear{Ge, Lin, and Yu}{Ge et~al\mbox{.}}{2019}]%
	{complementarypart2019}
	\bibfield{author}{\bibinfo{person}{Weifeng Ge} et al.} \bibinfo{year}{2019}\natexlab{}.
	\newblock \showarticletitle{Weakly supervised complementary parts models for
		fine-grained image classification from the bottom up}. In
	\bibinfo{booktitle}{\emph{CVPR}}. \bibinfo{pages}{3034--3043}.
	\newblock
	
	
	\bibitem[\protect\citeauthoryear{Goldberger and Ben-Reuven}{Goldberger and
		Ben-Reuven}{2017}]%
	{goldberger2017}
	\bibfield{author}{\bibinfo{person}{Jacob Goldberger} et al.} \bibinfo{year}{2017}\natexlab{}.
	\newblock \showarticletitle{Training deep neural-networks using a noise
		adaptation layer}. In \bibinfo{booktitle}{\emph{ICLR}}.
	\newblock
	
	
	\bibitem[\protect\citeauthoryear{Han, Yao, Yu, Niu, Xu, Hu, Tsang, and
		Sugiyama}{Han et~al\mbox{.}}{2018}]%
	{coteaching}
	\bibfield{author}{\bibinfo{person}{Bo Han} et al.}
	\bibinfo{year}{2018}\natexlab{}.
	\newblock \showarticletitle{Co-teaching: Robust training of deep neural
		networks with extremely noisy labels}. In \bibinfo{booktitle}{\emph{NIPS}}. \bibinfo{pages}{8527--8537}.
	\newblock
	
	
	\bibitem[\protect\citeauthoryear{He, Zhang, Ren, and Sun}{He
		et~al\mbox{.}}{2016}]%
	{resnet}
	\bibfield{author}{\bibinfo{person}{Kaiming He} et al.}
	\bibinfo{year}{2016}\natexlab{}.
	\newblock \showarticletitle{Deep residual learning for image recognition}. In
	\bibinfo{booktitle}{\emph{CVPR}}. \bibinfo{pages}{770--778}.
	\newblock
	
	
	\bibitem[\protect\citeauthoryear{Hua, Shen, Zhang, and Tang}{Hua
		et~al\mbox{.}}{2016}]%
	{2016domain}
	\bibfield{author}{\bibinfo{person}{Xian-sheng Hua} et al.}
	\bibinfo{year}{2016}\natexlab{}.
	\newblock \showarticletitle{A domain robust approach for image dataset
		construction}. In \bibinfo{booktitle}{\emph{ACM MM}}. \bibinfo{pages}{212--216}.
	\newblock
	
	
	\bibitem[\protect\citeauthoryear{Huang, Xu, Tao, and Zhang}{Huang
		et~al\mbox{.}}{2016}]%
	{huang2016}
	\bibfield{author}{\bibinfo{person}{Shaoli Huang} et al.}
	\bibinfo{year}{2016}\natexlab{}.
	\newblock \showarticletitle{Part-stacked cnn for fine-grained visual
		categorization}. In \bibinfo{booktitle}{\emph{CVPR}}. \bibinfo{pages}{1173--1182}.
	\newblock
	
	
	\bibitem[\protect\citeauthoryear{Jiang, Zhou, Leung, Li, and Fei-Fei}{Jiang
		et~al\mbox{.}}{2017}]%
	{mentornet}
	\bibfield{author}{\bibinfo{person}{Lu Jiang} et al.} \bibinfo{year}{2017}\natexlab{}.
	\newblock \showarticletitle{Mentornet: Learning data-driven curriculum for very
		deep neural networks on corrupted labels}. In
	\bibinfo{booktitle}{\emph{ICML}}.
	\newblock
	
	
	\bibitem[\protect\citeauthoryear{Kong and Fowlkes}{Kong and Fowlkes}{2017}]%
	{kong2017}
	\bibfield{author}{\bibinfo{person}{Shu Kong} et al.} \bibinfo{year}{2017}\natexlab{}.
	\newblock \showarticletitle{Low-rank bilinear pooling for fine-grained
		classification}. In \bibinfo{booktitle}{\emph{CVPR}}. \bibinfo{pages}{365--374}.
	\newblock
	
	
	\bibitem[\protect\citeauthoryear{Krause, Stark, Deng, and Fei-Fei}{Krause
		et~al\mbox{.}}{2013}]%
	{stanford-cars}
	\bibfield{author}{\bibinfo{person}{Jonathan Krause} et al.}
	\bibinfo{year}{2013}\natexlab{}.
	\newblock \showarticletitle{3d object representations for fine-grained
		categorization}. In \bibinfo{booktitle}{\emph{CVPRW}}.
	\bibinfo{pages}{554--561}.
	\newblock
	
	
	\bibitem[\protect\citeauthoryear{Krizhevsky and Hinton}{Krizhevsky and
		Hinton}{2009}]%
	{krizhevsky2009}
	\bibfield{author}{\bibinfo{person}{Alex Krizhevsky} et al.} \bibinfo{year}{2009}\natexlab{}.
	\newblock \showarticletitle{Learning multiple layers of features from tiny
		images}.
	\newblock \bibinfo{journal}{\emph{Technical report, University of Toronto}}
	\bibinfo{volume}{1}, \bibinfo{number}{4} (\bibinfo{year}{2009}),
	\bibinfo{pages}{7}.
	\newblock
	
	
	\bibitem[\protect\citeauthoryear{Kumar, Packer, and Koller}{Kumar
		et~al\mbox{.}}{2010}]%
	{kumar2010}
	\bibfield{author}{\bibinfo{person}{M~Pawan Kumar} et al.}
	\bibinfo{year}{2010}\natexlab{}.
	\newblock \showarticletitle{Self-paced learning for latent variable models}. In
	\bibinfo{booktitle}{\emph{NIPS}}. \bibinfo{pages}{1189--1197}.
	\newblock
	
	
	\bibitem[\protect\citeauthoryear{Lam, Mahasseni, and Todorovic}{Lam
		et~al\mbox{.}}{2017}]%
	{hsnet2017}
	\bibfield{author}{\bibinfo{person}{Michael Lam} et al.}
	\bibinfo{year}{2017}\natexlab{}.
	\newblock \showarticletitle{Fine-grained recognition as hsnet search for
		informative image parts}. In \bibinfo{booktitle}{\emph{CVPR}}. \bibinfo{pages}{2520--2529}.
	\newblock
	
	
	\bibitem[\protect\citeauthoryear{Li, Xie, Wang, and Gao}{Li
		et~al\mbox{.}}{2018}]%
	{li2018}
	\bibfield{author}{\bibinfo{person}{Peihua Li} et al.}
	\bibinfo{year}{2018}\natexlab{}.
	\newblock \showarticletitle{Towards faster training of global covariance
		pooling networks by iterative matrix square root normalization}. In
	\bibinfo{booktitle}{\emph{CVPR}}. \bibinfo{pages}{947--955}.
	\newblock
	
	
	\bibitem[\protect\citeauthoryear{Lin, Goyal, Girshick, He, and Doll{\'a}r}{Lin
		et~al\mbox{.}}{2017a}]%
	{focalloss2017}
	\bibfield{author}{\bibinfo{person}{Tsung-Yi Lin} et al.} \bibinfo{year}{2017}\natexlab{a}.
	\newblock \showarticletitle{Focal loss for dense object detection}. In
	\bibinfo{booktitle}{\emph{ICCV}}.
	\bibinfo{pages}{2980--2988}.
	\newblock
	
	
	\bibitem[\protect\citeauthoryear{Lin and Maji}{Lin and Maji}{2017}]%
	{linmaji2017}
	\bibfield{author}{\bibinfo{person}{Tsung-Yu Lin} et al.} \bibinfo{year}{2017}\natexlab{}.
	\newblock \showarticletitle{Improved bilinear pooling with cnns}. In
	\bibinfo{booktitle}{\emph{BMVC}}.
	\bibinfo{pages}{117.1--117.12}.
	\newblock
	
	
	\bibitem[\protect\citeauthoryear{Lin, RoyChowdhury, and Maji}{Lin
		et~al\mbox{.}}{2017b}]%
	{lin2017}
	\bibfield{author}{\bibinfo{person}{Tsung-Yu Lin} et al.}
	\bibinfo{year}{2017}\natexlab{b}.
	\newblock \showarticletitle{Bilinear convolutional neural networks for
		fine-grained visual recognition}.
	\newblock  \bibinfo{volume}{40}, \bibinfo{number}{6} (\bibinfo{year}{2017}),
	\bibinfo{pages}{1309--1322}.
	\newblock
	
	
	\bibitem[\protect\citeauthoryear{Lu, Liu, Yao, Tao, Tang, and Lu}{Lu
		et~al\mbox{.}}{2020}]%
	{lu2020hsi}
	\bibfield{author}{\bibinfo{person}{Jiarou Lu} et al.}
	\bibinfo{year}{2020}\natexlab{}.
	\newblock \showarticletitle{Hsi Road: A Hyper Spectral Image Dataset For Road
		Segmentation}. In \bibinfo{booktitle}{\emph{ICME}}. \bibinfo{pages}{1--6}.
	\newblock
	
	
	\bibitem[\protect\citeauthoryear{Luo, Lin, Liu, Liu, Tang, and Yao}{Luo
		et~al\mbox{.}}{2019}]%
	{luo2019segeqa}
	\bibfield{author}{\bibinfo{person}{Haonan Luo} et al.}
	\bibinfo{year}{2019}\natexlab{}.
	\newblock \showarticletitle{Segeqa: Video segmentation based visual attention
		for embodied question answering}. In \bibinfo{booktitle}{\emph{ICCV}}. \bibinfo{pages}{9667--9676}.
	\newblock
	
	
	\bibitem[\protect\citeauthoryear{Maji, Rahtu, Kannala, Blaschko, and
		Vedaldi}{Maji et~al\mbox{.}}{2013}]%
	{fgvc-aircraft}
	\bibfield{author}{\bibinfo{person}{Subhransu Maji} et al.} \bibinfo{year}{2013}\natexlab{}.
	\newblock \showarticletitle{Fine-grained visual classification of aircraft}.
	\newblock \bibinfo{journal}{\emph{arXiv:1306.5151}} (\bibinfo{year}{2013}).
	\newblock
	
	
	\bibitem[\protect\citeauthoryear{Malach and Shalev-Shwartz}{Malach and
		Shalev-Shwartz}{2017}]%
	{decoupling}
	\bibfield{author}{\bibinfo{person}{Eran Malach} et al.} \bibinfo{year}{2017}\natexlab{}.
	\newblock \showarticletitle{Decoupling "when to update" from "how to update"}.
	In \bibinfo{booktitle}{\emph{NIPS}}. \bibinfo{pages}{960--970}.
	\newblock
	
	
	\bibitem[\protect\citeauthoryear{Niu, Veeraraghavan, and Sabharwal}{Niu
		et~al\mbox{.}}{2018}]%
	{niu2018webly}
	\bibfield{author}{\bibinfo{person}{Li Niu} et al.}
	\bibinfo{year}{2018}\natexlab{}.
	\newblock \showarticletitle{Webly supervised learning meets zero-shot learning:
		A hybrid approach for fine-grained classification}. In
	\bibinfo{booktitle}{\emph{CVPR}}. \bibinfo{pages}{7171--7180}.
	\newblock
	
	
	\bibitem[\protect\citeauthoryear{Patrini, Rozza, Krishna~Menon, Nock, and
		Qu}{Patrini et~al\mbox{.}}{2017}]%
	{fcorrection}
	\bibfield{author}{\bibinfo{person}{Giorgio Patrini} et al.}
	\bibinfo{year}{2017}\natexlab{}.
	\newblock \showarticletitle{Making deep neural networks robust to label noise:
		A loss correction approach}. In \bibinfo{booktitle}{\emph{CVPR}}. \bibinfo{pages}{1944--1952}.
	\newblock
	
	
	\bibitem[\protect\citeauthoryear{Reed, Lee, Anguelov, Szegedy, Erhan, and
		Rabinovich}{Reed et~al\mbox{.}}{2014}]%
	{bootstrap}
	\bibfield{author}{\bibinfo{person}{Scott Reed} et al.}
	\bibinfo{year}{2014}\natexlab{}.
	\newblock \showarticletitle{Training deep neural networks on noisy labels with
		bootstrapping}. In \bibinfo{booktitle}{\emph{ICLR}}. \bibinfo{pages}{1--11}.
	\newblock
	
	
	\bibitem[\protect\citeauthoryear{Ren, Zeng, Yang, and Urtasun}{Ren
		et~al\mbox{.}}{2018}]%
	{ren2018}
	\bibfield{author}{\bibinfo{person}{Mengye Ren} et al.}
	\bibinfo{year}{2018}\natexlab{}.
	\newblock \showarticletitle{Learning to reweight examples for robust deep
		learning}. In \bibinfo{booktitle}{\emph{ICML}}. \bibinfo{pages}{4334--4343}.
	\newblock
	
	
	\bibitem[\protect\citeauthoryear{Shrivastava, Gupta, and Girshick}{Shrivastava
		et~al\mbox{.}}{2016}]%
	{harddatamining2016}
	\bibfield{author}{\bibinfo{person}{Abhinav Shrivastava} et al.}
	\bibinfo{year}{2016}\natexlab{}.
	\newblock \showarticletitle{Training region-based object detectors with online
		hard example mining}. In \bibinfo{booktitle}{\emph{CVPR}}. \bibinfo{pages}{761--769}.
	\newblock
	
	
	\bibitem[\protect\citeauthoryear{Shu, Tang, Qi, Liu, and Yang}{Shu
		et~al\mbox{.}}{2019}]%
	{shu2019hierarchical}
	\bibfield{author}{\bibinfo{person}{Xiangbo Shu} et al.} \bibinfo{year}{2019}\natexlab{}.
	\newblock \showarticletitle{Hierarchical long short-term concurrent memory for
		human interaction recognition}.
	\newblock \bibinfo{journal}{\emph{TPAMI}} (\bibinfo{year}{2019}).
	\newblock
	
	
	\bibitem[\protect\citeauthoryear{Simonyan and Zisserman}{Simonyan and
		Zisserman}{2014}]%
	{vgg}
	\bibfield{author}{\bibinfo{person}{Karen Simonyan} et al.} \bibinfo{year}{2014}\natexlab{}.
	\newblock \showarticletitle{Very deep convolutional networks for large-scale
		image recognition}.
	\newblock \bibinfo{journal}{\emph{arXiv:1409.1556}} (\bibinfo{year}{2014}).
	\newblock
	
	
	\bibitem[\protect\citeauthoryear{Song, Kim, and Lee}{Song
		et~al\mbox{.}}{2019}]%
	{selfie2019}
	\bibfield{author}{\bibinfo{person}{Hwanjun Song} et al.} \bibinfo{year}{2019}\natexlab{}.
	\newblock \showarticletitle{SELFIE: Refurbishing unclean samples for robust
		deep learning}. In \bibinfo{booktitle}{\emph{ICML}}. \bibinfo{pages}{5907--5915}.
	\newblock
	
	
	\bibitem[\protect\citeauthoryear{Sun, Shen, Liu, and Wang}{Sun
		et~al\mbox{.}}{2019}]%
	{yao2019dynamically}
	\bibfield{author}{\bibinfo{person}{Zeren Sun} et al.}
	\bibinfo{year}{2019}\natexlab{}.
	\newblock \showarticletitle{Dynamically visual disambiguation of keyword-based
		image search}.
	\newblock \bibinfo{journal}{\emph{IJCAI}} (\bibinfo{year}{2019}), \bibinfo{pages}{996--1002}.
	\newblock
	
	
	\bibitem[\protect\citeauthoryear{Szegedy, Vanhoucke, Ioffe, Shlens, and
		Wojna}{Szegedy et~al\mbox{.}}{2016}]%
	{szegedy2016}
	\bibfield{author}{\bibinfo{person}{Christian Szegedy} et al.} \bibinfo{year}{2016}\natexlab{}.
	\newblock \showarticletitle{Rethinking the inception architecture for computer
		vision}. In \bibinfo{booktitle}{\emph{CVPR}}. \bibinfo{pages}{2818--2826}.
	\newblock
	
	
	\bibitem[\protect\citeauthoryear{Tang, Li, Lai, Zhang, Yan, et~al\mbox{.}}{Tang
		et~al\mbox{.}}{2017}]%
	{tang2017personalized}
	\bibfield{author}{\bibinfo{person}{Jinhui Tang} et al.}
	\bibinfo{year}{2017}\natexlab{}.
	\newblock \showarticletitle{Personalized age progression with bi-level aging
		dictionary learning}.
	\newblock \bibinfo{journal}{\emph{TPAMI}} \bibinfo{volume}{40}, \bibinfo{number}{4}
	(\bibinfo{year}{2017}), \bibinfo{pages}{905--917}.
	\newblock
	
	
	\bibitem[\protect\citeauthoryear{Van~Horn, Mac~Aodha, Song, Cui, Sun, Shepard,
		Adam, Perona, and Belongie}{Van~Horn et~al\mbox{.}}{2018}]%
	{inat17}
	\bibfield{author}{\bibinfo{person}{Grant Van~Horn} et al.} \bibinfo{year}{2018}\natexlab{}.
	\newblock \showarticletitle{The inaturalist species classification and
		detection dataset}. In \bibinfo{booktitle}{\emph{CVPR}}. \bibinfo{pages}{8769--8778}.
	\newblock
	
	
	\bibitem[\protect\citeauthoryear{Wah, Branson, Welinder, Perona, and
		Belongie}{Wah et~al\mbox{.}}{2011}]%
	{cub200-2011}
	\bibfield{author}{\bibinfo{person}{Catherine Wah} et al.} \bibinfo{year}{2011}\natexlab{}.
	\newblock \showarticletitle{The Caltech-UCSD Birds-200-2011 Dataset.}
	\newblock \bibinfo{journal}{\emph{CNS-TR-2011-001}} (\bibinfo{year}{2011}).
	\newblock
	
	
	\bibitem[\protect\citeauthoryear{Wei, Xie, Wu, and Shen}{Wei
		et~al\mbox{.}}{2018}]%
	{wei2018}
	\bibfield{author}{\bibinfo{person}{Xiu-Shen Wei} et al.}
	\bibinfo{year}{2018}\natexlab{}.
	\newblock \showarticletitle{Mask-CNN: Localizing parts and selecting
		descriptors for fine-grained bird species categorization}.
	\newblock \bibinfo{journal}{\emph{PR}}  \bibinfo{volume}{76}
	(\bibinfo{year}{2018}), \bibinfo{pages}{704--714}.
	\newblock
	
	
	\bibitem[\protect\citeauthoryear{Xie, Liu, Jin, Zhu, Zhang, Qin, Yao, and
		Shao}{Xie et~al\mbox{.}}{2019}]%
	{xie2019attentive}
	\bibfield{author}{\bibinfo{person}{Guo-Sen Xie} et al.}
	\bibinfo{year}{2019}\natexlab{}.
	\newblock \showarticletitle{Attentive region embedding network for zero-shot
		learning}. In \bibinfo{booktitle}{\emph{CVPR}}. \bibinfo{pages}{9384--9393}.
	\newblock
	
	
	\bibitem[\protect\citeauthoryear{Xie, Liu, Zhu, Zhao, Zhang, Yao, Qin, and
		Shao}{Xie et~al\mbox{.}}{2020}]%
	{xie2020eccv}
	\bibfield{author}{\bibinfo{person}{Guo-Sen Xie} et al.}
	\bibinfo{year}{2020}\natexlab{}.
	\newblock \showarticletitle{Region Graph Embedding Network for Zero-Shot
		Learning}. In \bibinfo{booktitle}{\emph{ECCV}}.
	\newblock
	
	
	\bibitem[\protect\citeauthoryear{Xu, Huang, Zhang, and Tao}{Xu
		et~al\mbox{.}}{2016}]%
	{xu2018webly}
	\bibfield{author}{\bibinfo{person}{Zhe Xu} et al.}
	\bibinfo{year}{2016}\natexlab{}.
	\newblock \showarticletitle{Webly-supervised fine-grained visual categorization
		via deep domain adaptation}.
	\newblock  \bibinfo{volume}{40}, \bibinfo{number}{5} (\bibinfo{year}{2016}),
	\bibinfo{pages}{1100--1113}.
	\newblock
	
	
	\bibitem[\protect\citeauthoryear{Yang, Sun, Lai, Zheng, and Cheng}{Yang
		et~al\mbox{.}}{2018}]%
	{yang2018recognition}
	\bibfield{author}{\bibinfo{person}{Jufeng Yang} et al.} \bibinfo{year}{2018}\natexlab{}.
	\newblock \showarticletitle{Recognition from web data: A progressive filtering
		approach}.
	\newblock  \bibinfo{volume}{27}, \bibinfo{number}{11} (\bibinfo{year}{2018}),
	\bibinfo{pages}{5303--5315}.
	\newblock
	
	
	\bibitem[\protect\citeauthoryear{Yao, Zhang, Zhang, Li, and Tian}{Yao
		et~al\mbox{.}}{2016}]%
	{coarse-to-fine}
	\bibfield{author}{\bibinfo{person}{Hantao Yao} et al.} \bibinfo{year}{2016}\natexlab{}.
	\newblock \showarticletitle{Coarse-to-fine description for fine-grained visual
		categorization}.
	\newblock  \bibinfo{volume}{25}, \bibinfo{number}{10} (\bibinfo{year}{2016}),
	\bibinfo{pages}{4858--4872}.
	\newblock
	
	
	\bibitem[\protect\citeauthoryear{Yao, Shen, Xie, Liu, Zhu, Zhang, and Shen}{Yao
		et~al\mbox{.}}{2020}]%
	{yao2020exploiting}
	\bibfield{author}{\bibinfo{person}{Yazhou Yao} et al.}
	\bibinfo{year}{2020}\natexlab{}.
	\newblock \showarticletitle{Exploiting web images for multi-output
		classification: From category to subcategories}.
	\newblock \bibinfo{journal}{\emph{TNNLS}} \bibinfo{volume}{31}, \bibinfo{number}{7}
	(\bibinfo{year}{2020}), \bibinfo{pages}{2348--2360}.
	\newblock
	
	
	\bibitem[\protect\citeauthoryear{Yao, Shen, Zhang, Liu, Tang, and Shao}{Yao
		et~al\mbox{.}}{2018a}]%
	{yao2018extracting}
	\bibfield{author}{\bibinfo{person}{Yazhou Yao} et al.}
	\bibinfo{year}{2019}\natexlab{a}.
	\newblock \showarticletitle{Extracting multiple visual senses for web
		learning}.
	\newblock \bibinfo{journal}{\emph{TMM}}
	\bibinfo{volume}{21}, \bibinfo{number}{1} (\bibinfo{year}{2019}),
	\bibinfo{pages}{184--196}.
	\newblock
	
	
	\bibitem[\protect\citeauthoryear{Yao, Shen, Zhang, Liu, Tang, and Shao}{Yao
		et~al\mbox{.}}{2018b}]%
	{yao2018tip}
	\bibfield{author}{\bibinfo{person}{Yazhou Yao} et al.}
	\bibinfo{year}{2019}\natexlab{b}.
	\newblock \showarticletitle{Extracting privileged information for enhancing
		classifier learning}.
	\newblock \bibinfo{journal}{\emph{TIP}}
	\bibinfo{volume}{28}, \bibinfo{number}{1} (\bibinfo{year}{2019}),
	\bibinfo{pages}{436--450}.
	\newblock
	
	
	\bibitem[\protect\citeauthoryear{Yao, Zhang, Shen, Hua, Xu, and Tang}{Yao
		et~al\mbox{.}}{2017}]%
	{yao2017exploiting}
	\bibfield{author}{\bibinfo{person}{Yazhou Yao} et al.}
	\bibinfo{year}{2017}\natexlab{}.
	\newblock \showarticletitle{Exploiting web images for dataset construction: A
		domain robust approach}.
	\newblock \bibinfo{journal}{\emph{TMM}}
	\bibinfo{volume}{19}, \bibinfo{number}{8} (\bibinfo{year}{2017}),
	\bibinfo{pages}{1771--1784}.
	\newblock
	
	
	\bibitem[\protect\citeauthoryear{Yao, Zhang, Shen, Liu, Zhu, Zhang, and
		Shen}{Yao et~al\mbox{.}}{2020}]%
	{yao2019towards}
	\bibfield{author}{\bibinfo{person}{Yazhou Yao} et al.} \bibinfo{year}{2019}\natexlab{}.
	\newblock \showarticletitle{Towards automatic construction of diverse,
		high-quality image datasets}.
	\newblock \bibinfo{journal}{\emph{TKDE}} \bibinfo{volume}{32}, \bibinfo{number}{6}
	(\bibinfo{year}{2020}), \bibinfo{pages}{1199--1211}.
	\newblock
	
	
	\bibitem[\protect\citeauthoryear{Yi and Wu}{Yi and Wu}{2019}]%
	{pencil2019}
	\bibfield{author}{\bibinfo{person}{Kun Yi} et al.}
	\bibinfo{year}{2019}\natexlab{}.
	\newblock \showarticletitle{Probabilistic end-to-end noise correction for
		learning with noisy labels}. In \bibinfo{booktitle}{\emph{CVPR}}. \bibinfo{pages}{7017--7025}.
	\newblock
	
	
	\bibitem[\protect\citeauthoryear{Zhang, Bengio, Hardt, Recht, and
		Vinyals}{Zhang et~al\mbox{.}}{2017a}]%
	{zhang2016understanding}
	\bibfield{author}{\bibinfo{person}{Chiyuan Zhang} et al.} \bibinfo{year}{2017}\natexlab{a}.
	\newblock \showarticletitle{Understanding deep learning requires rethinking
		generalization}. In \bibinfo{booktitle}{\emph{ICLR}}.
	\newblock
	
	
	\bibitem[\protect\citeauthoryear{Zhang, Yao, Liu, Xie, Shu, Zhou, Zhang, Shen,
		and Tang}{Zhang et~al\mbox{.}}{2020a}]%
	{aaai20}
	\bibfield{author}{\bibinfo{person}{Chuanyi Zhang} et al.} \bibinfo{year}{2020}\natexlab{a}.
	\newblock \showarticletitle{Web-Supervised Network with Softly Update-Drop
		Training for Fine-Grained Visual Classification}. In
	\bibinfo{booktitle}{\emph{AAAI}}.
	\bibinfo{pages}{12781--12788}.
	\newblock
	
	
	\bibitem[\protect\citeauthoryear{Zhang, Yao, Zhang, Chen, and et~al.}{Zhang
		et~al\mbox{.}}{2020b}]%
	{zhang2020web}
	\bibfield{author}{\bibinfo{person}{Chuanyi Zhang} et al.} \bibinfo{year}{2020}\natexlab{b}.
	\newblock \showarticletitle{Web-Supervised Network for Fine-Grained Visual
		Classification}. In \bibinfo{booktitle}{\emph{ICME}}. \bibinfo{pages}{1--6}.
	\newblock
	
	
	\bibitem[\protect\citeauthoryear{Zhang, Shen, Hua, Xu, and Tang}{Zhang
		et~al\mbox{.}}{2016}]%
	{2016automatic}
	\bibfield{author}{\bibinfo{person}{Fumin Shen} et al.} \bibinfo{year}{2016}\natexlab{}.
	\newblock \showarticletitle{Automatic image dataset construction with multiple
		textual metadata}. In \bibinfo{booktitle}{\emph{ICME}}. \bibinfo{pages}{1--6}.
	\newblock
	
	
	\bibitem[\protect\citeauthoryear{Zhang, Shen, Hua, Xu, and Tang}{Zhang
		et~al\mbox{.}}{2017b}]%
	{2017new}
	\bibfield{author}{\bibinfo{person}{Jian Zhang} et al.} \bibinfo{year}{2017}\natexlab{b}.
	\newblock \showarticletitle{A new web-supervised method for image dataset
		constructions}.
	\newblock \bibinfo{journal}{\emph{Neurocomputing}}  \bibinfo{volume}{236}
	(\bibinfo{year}{2017}), \bibinfo{pages}{23--31}.
	\newblock
	
	
	\bibitem[\protect\citeauthoryear{Zhang, Shen, Yang, Huang, and Tang}{Zhang
		et~al\mbox{.}}{2018}]%
	{2018discovering}
	\bibfield{author}{\bibinfo{person}{Jian Zhang} et al.} \bibinfo{year}{2018}\natexlab{}.
	\newblock \showarticletitle{Discovering and distinguishing multiple visual
		senses for polysemous words}. In \bibinfo{booktitle}{\emph{AAAI}}. \bibinfo{pages}{523--530}.
	\newblock
	
	
	\bibitem[\protect\citeauthoryear{Zheng, Fu, Zha, and Luo}{Zheng
		et~al\mbox{.}}{2019}]%
	{trilinear2019}
	\bibfield{author}{\bibinfo{person}{Heliang Zheng} et al.}
	\bibinfo{year}{2019}\natexlab{}.
	\newblock \showarticletitle{Looking for the devil in the details: Learning
		trilinear attention sampling network for fine-grained image recognition}. In
	\bibinfo{booktitle}{\emph{CVPR}}. \bibinfo{pages}{5012--5021}.
	\newblock
	
	
	\bibitem[\protect\citeauthoryear{Zhou, Khosla, Lapedriza, Torralba, and
		Oliva}{Zhou et~al\mbox{.}}{2016}]%
	{pcaduplicateremoval2016}
	\bibfield{author}{\bibinfo{person}{Bolei Zhou} et al.}
	\bibinfo{year}{2016}\natexlab{}.
	\newblock \showarticletitle{Places: An image database for deep scene
		understanding}.
	\newblock \bibinfo{journal}{\emph{arXiv:1610.02055}} (\bibinfo{year}{2016}).
	\newblock
	
	
\end{thebibliography}
\end{document}